\documentclass{article}

\usepackage{arxiv}

\usepackage[utf8]{inputenc} 
\usepackage[T1]{fontenc}    
\usepackage{hyperref}       
\usepackage{url}            
\usepackage{booktabs}       
\usepackage{amsfonts}       
\usepackage{nicefrac}       
\usepackage{microtype}      
\usepackage{lipsum}
\usepackage{amsmath}
\usepackage{amssymb}
\usepackage{graphicx}
\usepackage{multirow}
\usepackage{multicol}
\usepackage{authblk}
\usepackage{todonotes}
\usepackage{cite}
\usepackage{float}
\usepackage{pdflscape}
\usepackage{algorithmic}
\usepackage{algorithm}

\DeclareMathOperator*{\argmin}{arg\,min}

\renewcommand\vec{\mathbf}
\renewcommand\ref{\href}
\newcommand{\hospace}{1pt}
\newcommand{\real}{{\rm I\!R}}

\newcommand{\loss}{\mathcal{L}}
\newcommand{\ucishort}{IHEPC}
\newcommand{\uci}{Individual household electric power consumption data set }
\newcommand{\gcshort}{GEFCom2014}

\newcommand{\inlen}{n_T}
\newcommand{\outlen}{n_O}

\newcommand{\transpose}[1]{#1^{\text{T}}}
\newcommand{\specialcell}[2][c]{%
  \begin{tabular}[#1]{@{}c@{}}#2\end{tabular}}
\renewcommand{\t}[1]{[#1]}

\title{Deep Learning for Time Series Forecasting: \\ The Electric Load Case}

\begin{document}
\maketitle

\vbox{%
\hsize\textwidth
\linewidth\hsize
\centering
\begin{tabular}[t]{c}
    \textbf{Alberto Gasparin}\\
    Faculty of Informatics \\
    Università della Svizzera Italiana \\
    6900 Lugano, Switzerland \\
    \texttt{alberto.gasparin@usi.ch} \\
\end{tabular}
\hfil\linebreak[0]\hfil%
\begin{tabular}[t]{c}
\textbf{Slobodan Lukovic} \\ 
Faculty of Informatics \\
Università della Svizzera Italiana \\
6900 Lugano, Switzerland \\
\texttt{slobodan.lukovic@usi.ch} \\
\end{tabular}
\hfil\linebreak[0]\hfil%
\begin{tabular}[t]{c}
\textbf{Cesare Alippi} \\ 
    Dept. of Electronics, Information, and Bioengineering \\
    Politecnico di Milano \\
    20133 Milan, Italy, \\
    Faculty of Informatics \\
    Università della Svizzera Italiana \\
    6900 Lugano, Switzerland \\
    \texttt{cesare.alippi@polimi.it} \\
\end{tabular}
\vskip 0.4in
}%

\begin{abstract}
Management and efficient operations in critical infrastructure such as Smart Grids take huge advantage of accurate power load forecasting which, due to its nonlinear nature, remains a challenging task.
Recently, deep learning has emerged in the machine learning field achieving impressive performance in a vast range of tasks, from image classification to machine translation. 
Applications of deep learning models to the electric load forecasting problem are gaining interest among researchers as well as the industry, but a comprehensive and sound comparison among different architectures is not yet available in the literature. 
This work aims at filling the gap by reviewing and experimentally evaluating on two real-world datasets the most recent trends in electric load forecasting, by contrasting deep learning architectures on short term forecast (one day ahead prediction).
Specifically, we focus on feedforward and recurrent neural networks, sequence to sequence models and temporal convolutional neural networks along with architectural variants, which are known in the signal processing community but are novel to the load forecasting one.
\end{abstract}


\section{Introduction}
Smart grids aim at creating automated and efficient energy delivery networks which improve power delivery reliability and quality, along with network security, energy efficiency, and  demand-side management aspects \cite{SmartGrid}.
Modern power distribution systems are supported by advanced monitoring infrastructures that produce immense amount of data, thus enabling fine grained analytics and improved forecasting performance.
In particular, electric load forecasting emerges as a critical task in the energy field, as it enables useful support for decision making, supporting optimal pricing strategies, seamless integration of renewables and maintenance cost reductions.
Load forecasting is carried out at different time horizons, ranging from milliseconds to years, depending on the specific problem at hand. 

In this work we focus on the day-ahead prediction problem also referred in the literature as \textit{short term load forecasting} (STLF) \cite{STLFClass}. Since deregulation of electric energy distribution and wide adoption of renewables strongly affects daily market prices, STLF emerges to be of fundamental importance for efficient power supply \cite{ANNReview}. 
Furthermore, we differentiate forecasting on the granularity level at which it is applied. For instance, in individual household scenario, load prediction is rather difficult as power consumption patterns are highly volatile. On the contrary, aggregated load consumption i.e., that associated with a neighborhood, a region, or even an entire state, is normally easier to predict as the resulting signal exhibit slower dynamics.

Historical power loads are time-series affected by several external time-variant factors, such as weather conditions, human activities, temporal and seasonal characteristics that make their predictions a challenging problem.
A large variety of prediction methods has been proposed for the electric load forecasting over the years and, only the most relevant ones are reviewed in this section.
Autoregressive moving average models (ARMA) were among the first model families used in short-term load forecasting \cite{ARMA1, ARMA2}. Soon they were replaced by ARIMA and seasonal ARIMA models \cite{TSA} to cope with time variance often exhibited by load profiles. In order to include exogenous variables like temperature into the forecasting method, model families were extended to ARMAX \cite{ARMAX, ARMAX2} and ARIMAX\cite{ARIMAX}.
The main shortcoming of these system identification families is the linearity assumption for the system being observed, hypothesis that does not generally hold. 
In order to solve this limitation, nonlinear models like Feed Forward Neural Networks were proposed and became attractive for those scenarios exhibiting significant nonlinearity, as in load forecasting tasks \cite{LeeANN, ParkANN, SrinANN, DrzegaANN, ANNReview}.
The intrinsic sequential nature of time series data was then exploited by considering sophisticated techniques ranging from advanced feed forward architecture with residual connections \cite{ResNet_STLM} to convolutional approaches \cite{DeepEnergy,MarinoCNN} and Recurrent Neural Networks \cite{ElmannSTLF, RNN_Bianchi} along with their many variants such as Echo-state Network \cite{BianchiPCAESN, Mocanu, RNN_Bianchi}, Long-Short Term Memory \cite{ESTLF_LSTM, STLF_LSTM, GA_LSTM, RNN_Bianchi} and Gated Recurrent Unit \cite{GRU_STLF, RNN_Bianchi}.
Moreover, some hybrid architectures have also been proposed aiming to capture the temporal dependencies in the data with recurrent networks while performing a more general feature extraction operation with convolutional layers \cite{CNN_RNN, CNN_LSTM}. 

\begin{table*}
\centering
    \begin{tabular}{ccccc}
        \toprule
        \textbf{Reference}    & \textbf{Predictive Family of Models} & \textbf{Time Horizon} & \textbf{Exogenous Variables}  & \textbf{Dataset (Location)} \\
        \toprule
        \cite{RNN_Bianchi} & LSTM, GRU, ERNN, NARX, ESN                   & D                & -              & Rome, Italy \\
        \cite{RNN_Bianchi} & LSTM, GRU, ERNN, NARX, ESN                   & D                & T              & New England \cite{GEFCom2012} \\
        \cite{ElmannPalermo} & ERNN & H & T, H, P, other$^{*}$ & Palermo, Italy \\
        \cite{ElmannSTLF} & ERNN      & H                & T, W, H                 &  Hubli, India \\ 
        \cite{Mocanu} & ESN & 15min to 1Y & - & Sceaux, France \cite{UCI} \\
        \cite{ESTLF_LSTM} & LSTM, NARX      & D                & -                 & Unknown \\
        \cite{STLF_LSTM}  & LSTM      & D(?)                & C, TI                                                           & Australia \cite{SGSC}  \\
        \cite{GA_LSTM}    & LSTM      & 2W to 4M            & T, W, H, C, TI
        & France  \\
        \cite{PowerLSTM}  & LSTM      & 2D                  & T, P, H, C, TI                                                       & Unknown \cite{UMData}\\
        \midrule
        \cite{DLMarino}  & LSTM, seq2seq-LSTM                   & 60 H                & C, TI              & Sceaux, France \cite{UCI} \\
        \cite{Wilms}     & LSTM, seq2seq-LSTM                   & 12 H                & T, C, TI              & New England \cite{GEFCom2014} \\
        \cite{GRU_STLF} & GRU & D & T, C, other$^{**}$ & Dongguan, China\\
        \midrule
        \cite{DeepEnergy} & CNN                   & D                & C, TI              & USA \\
        \cite{MarinoCNN}  & CNN                   & D                & C, TI              & Sceaux, France \\
        \cite{CNN_RNN}    & CNN + LSTM            & D                & T, C, TI              & North-China \\
        \cite{CNN_LSTM}   & CNN + LSTM            & D                & -                            & North-Italy\\
        \bottomrule
    \end{tabular}
\caption{\textit{Time Horizon:} H(hour), D (day), W(week), M(month), Y(year), ?(Not explicitly stated, thus, inferred from text) \textit{Exogenous variables:} T (temperature), W (wind speed), H (humidity), P (pressure), C (calendar including date and holidays information), TI (time), $^{*}$ other input features were created for this dataset, $^{**}$ categorical weather information is used (e.g., sunny, cloudy), \textit{Dataset}: the data source, a link is provided whenever available.}
\label{tab:sota}
\end{table*}

Different reviews address the load forecasting topic by means of (not necessarily deep) neural networks. In \cite{STFL_DL_Survey} the authors focus on the use of some deep learning architectures for load forecasting. However, this review lacks a comprehensive comparative study of performance verified on common load forecasting benchmarks. 
The absence of valid cost-performance metric does not allow the report to make conclusive statements.
In \cite{RNN_Bianchi} an exhaustive overview of recurrent neural networks for short term load forecasting is presented. The very detailed work considers one layer (not deep) recurrent networks only.
A comprehensive summary of the most relevant researches dealing with STLF employing recurrent neural networks, convolutional neural networks and seq2seq models is presented in Table \ref{tab:sota}. It emerges that most of the works have been performed on different datasets, making it rather difficult - if not impossible - to asses their absolute performance and, consequently, recommend the best state-of-the-art solutions for load forecast.

In this survey we consider the most relevant -and recent- deep  architectures and contrast them in terms of performance accuracy on open-source benchmarks. 
The considered architectures include recurrent neural networks, sequence to sequence models and temporal convolutional neural networks. 
The experimental comparison is performed on two different real-world datasets which are representatives of two distinct scenarios. The first one considers power consumption at an individual household level with a signal characterized by high frequency components while the second one takes into account aggregation of several consumers.
Our contributions consist in:
\begin{itemize}
    \item A comprehensive review. The survey provides a comprehensive investigation of deep learning architectures known to the smart grid literature as well as novel recent ones suitable for electric load forecasting. 
    \item A multi-step prediction strategy comparison for recurrent neural networks: we study and compare how different prediction strategies can be applied to recurrent neural networks. To the best of our knowledge this work has not been done yet for deep recurrent neural networks.
    \item A relevant performance assessment. To the best of our knowledge, the present work provides the first systematic experimental comparison of the most relevant deep learning architectures for the electric load forecasting problems of individual and aggregated electric demand. It should be noted that envisaged architectures are domain independent and, as such, can be applied in different forecasting scenarios.
\end{itemize}
The rest of this paper is organized as follows. \\
In Section \ref{sec:notation} we formally introduce the forecasting problems along with the notation that will be used in this work.
In Section \ref{sec:fnn} we introduce Feed Forward Neural Networks (FNNs) and the main concepts relevant to the learning task. We also provide a short review of the literature regarding the use of FNNs for the load forecasting problem.\\
In Section \ref{sec:rnn} we provide a general overview of Recurrent Neural Networks (RNNs) and their most advanced architectures: Long Short-Term Memory and Gated Recurrent Unit networks. \\
In Section \ref{sec:s2s} Sequence To Sequence architectures (seq2seq) are discussed as a general improvement over recurrent neural networks. We present both, simple and advanced models built on the sequence to sequence paradigm. \\
In Section \ref{sec:cnn} Convolutional Neural Networks are introduced and one of their most recent variant, the temporal convolutional network (TCN), is presented as the state-of-the-art method for univariate time-series prediction. \\
In Section \ref{sec:experiments} the real-world datasets used for models comparison are presented. For each dataset, we provide a description of the preprocessing operations and the techniques that have been used to validate the models performance. \\ 
Finally, In Section \ref{sec:conclusions} we draw conclusions based on the performed assessments.

\section{Problem Description}
\label{sec:notation}
\begin{figure*}[ht]
    \includegraphics[width=\linewidth]{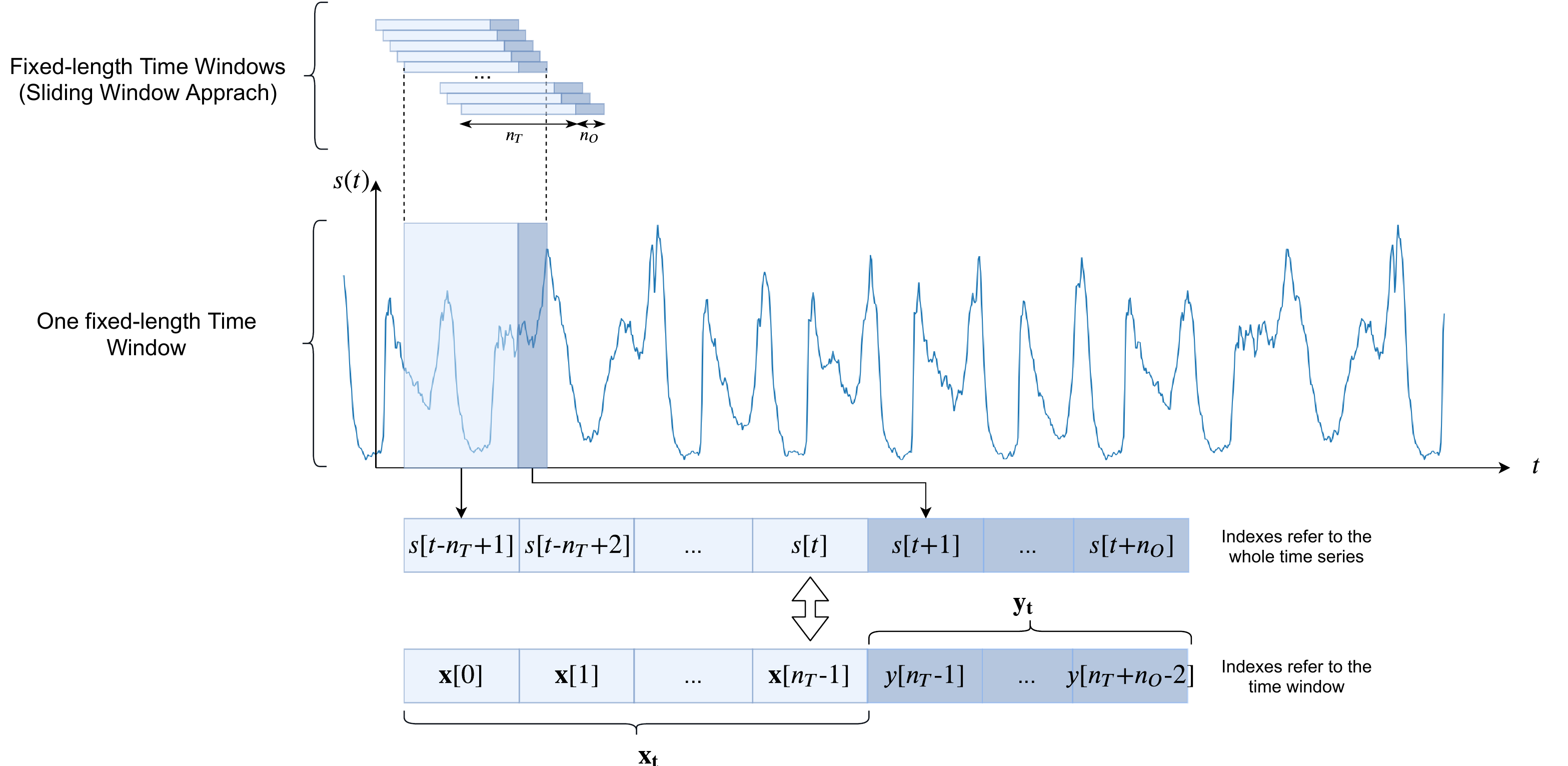}
\caption{A sliding windowed approach is used to frame the forecasting problem into a supervised machine learning problem. The target signal $\vec{s}$ is split in multiple input output pairs $(\vec{x_t}, \vec{y_t}) \forall t \in \{\inlen,\inlen+1, \dots, T-\outlen\}$ }
\label{fig:notation}
\end{figure*}

In basic multi-step ahead electric load forecasting a univariate time series $\vec{s} = [s\t{0}, s\t{1} \dots, s\t{T}]$ that spans through several years is given. In this work, input data are presented to the different predictive families of models as a regressor vector composed of fixed time-lagged data associated with a window size of length $\inlen$ which slides over the time series. 
Given this fixed length view of past values, a predictor $f$ aims at forecasting the next $\outlen$ values of the time series.
In this work the forecasting problem is studied as a supervised learning problem.
As such, given the input vector at discrete time $t$ defined as $\vec{x}_t = [s[t-\inlen+1], \dots, s[t]] \in \real^{\inlen}$, the forecasting problem requires to infer the next $\outlen$ measurements $\vec{y_t} = [s\t{t+1}, \dots, s\t{t+\outlen}] \in \real^{\outlen}$ or a subset of.
To ease the notation we express the input and output vectors in the reference system of the time window instead of the time series one. 
By following this approach, the input vector at discrete time $t$ becomes $\vec{x}_t = [x_t[0], \dots, x_t[\inlen-1]] \in \real^{\inlen}, x_t\t{i} = s[i+1+t-\inlen]$ and the corresponding output vector is $\vec{y_t} = [y_t\t{\inlen-1}, \dots, y_t\t{\inlen+\outlen-2}] \in \real^{\outlen}$. $y_t$ characterizes the real output values  defined as $y_t\t{t} = x_t\t{t+1}, \quad \forall t \in T$. Similarly, we denote as $\vec{\hat{y}_t}=f(\vec{x_t}; \vec{\hat{\Theta}}) \in \real^{\outlen}$, the prediction vector provided by a predictive model $f$ whose parameters vector $\vec{\Theta}$ has been estimated by optimizing a performance function.

Without loss of generality, in the remaining of the paper, we drop the subscript $t$ from the inner elements of $\vec{x_t}$ and $\vec{y_t}$. 
The introduced notation, along with the sliding window approach, is depicted in Figure \ref{fig:notation}.

In certain applications we will additionally be provided with $d-1$ exogenous variables (e.g., the temperatures) each of which representing a univariate time series aligned in time with the data of electricity demand. In this scenario the components of the regressor vector become vectors, i.e., $\vec{x_t} = [\vec{x}\t{0}, \dots, \vec{x}\t{\inlen-1}] \in \real^{\inlen \times d}$.
Indeed, each element of the input sequence is represented as $\vec{x}\t{t} = [x\t{t}, z_0\t{t}, \dots, z_{d-2}\t{t}] \in \real^{d}$ where $x\t{t} \in \real$ is the scalar load measurement at time $t$, while $z_k\t{t} \in \real$ is the scalar value of the $k^{th}$ exogenous feature.

The nomenclature used in this work is given in Table \ref{tab:notation}.

\begin{table}[]
\centering
\begin{tabular}{c|c}
\toprule
\textbf{Notation} & \textbf{Description}                                           \\
\toprule
$\inlen$          & window size of the regressor vector \\
\hline
$\outlen$         & time horizon of the forecast                                     \\
\hline
$d-1$             & number of exogenous variable \\       
\hline
$u$                 & scalar value                                                         \\
\hline
$\vec{u}$         & vector/matrix                                                  \\
\hline
$\transpose{\vec{u}}$       & vector/matrix transposed                                       \\
\hline
$\odot$           & elementwise product                                            \\
\hline
*                 & convolution operator                                          \\
\hline
$*_d$             & dilated convolution operation                                  \\
\hline
$\ell$            & index for the $l^{th}$ layer                                   \\
\hline
$\vec{x_t}$       & \specialcell{regressor vector at discrete time $t$ \\ (reference system: time-series) \\ $\vec{x_t} \in \real^{\inlen}$ or $\vec{x_t} \in \real^{\inlen \times d}$} \\
\hline
$\vec{y_t}$       & \specialcell{true output for the input sequence at time $t$ \\ (reference system: time-series)\\ $\vec{y_t} \in \real^{\outlen}$} \\
\hline
$\vec{\hat{y}_t}$ & \specialcell{predicted output for the input sequence at time $t$ \\ (reference system: time-series) \\ $\vec{\hat{y}_t} \in \real^{\outlen}$}\\
\hline
$\vec{x}\t{t}$    & \specialcell{input vector of load \& other features at time $t$ \\(reference system: time window)\\ $\vec{x}\t{t} \in \real$ or $\vec{x}\t{t} \in \real^d$} \\
\hline
$y\t{t}$          & \specialcell{value of the load time-series at time $t+1$ \\ (reference system: time window) \\ $y\t{t} \in \real$} \\
\hline
$\vec{z_t}$       & \specialcell{exogenous features vector at time $t$ \\ (reference system: time series) \\ $\vec{z_t} \in \real^{\inlen \times {d-1}}$}\\
\hline
$\vec{z}\t{t}$    & \specialcell{exogenous features vector at time $t$ \\ (reference system:time window).\\ $\vec{z}\t{t} \in \real^{d-1}$} \\
\hline
$\vec{h}$         & hidden state vector  \\
\hline
$\vec{\Theta}$    & model's vector of parameters \\
\hline
$n_H$             & number of hidden neurons \\
\bottomrule
\end{tabular}
\caption{The nomenclature used in this work.}
\label{tab:notation}
\end{table}

\section{Feed Forward Neural Networks}
\label{sec:fnn}
Feed Forward Neural Networks (FNNs) are parametric model families characterized by the universal function approximation property\cite{UFA}.
Their computational architectures are composed of a layered structure consisting of three main building blocks: the input layer, the hidden layer(s) and the output layer.
The number of hidden layers ($L > 1$), determines the depth of the network, while the size of each layer, i.e., the number $n_{H,\ell}$ of hidden units of the $\ell-\text{th}$ layer defines its complexity in terms of neurons.
FNNs provide only direct forward connections between two consecutive layers, each connection associated with a trainable parameter; note that given the feedfoward nature of the computation no recursive feedback is allowed. 
More in detail, given a vector $\vec{x} \in \real^{\inlen}$ fed at the network input, the FNN's computation  can be expressed as:
\begin{align}
    & \vec{a}_{\ell} = \vec{W}_{\ell}^\text{T} \vec{h_{\ell - 1}} + \vec{b}_{\ell} ,\quad \ell = 1, ... L  \label{eq:affinetransform} \\
    & \vec{h}_{\ell} = \phi_\ell(\vec{a}_{\ell}) 
    \label{eq:activation}
\end{align}
where $\vec{h}_0 = \vec{x_t} \in \real^{\inlen}$ and $\vec{\hat{y}_t} = \vec{h}_{L} \in \real^{\outlen}$.

Each layer $\ell$ is characterized with its own parameters matrix $\vec{W}_{\ell} \in \real^{n_{H,\ell-1} \times n_{H,\ell}}$ and bias vector $\vec{b}_{\ell} \in \real^{n_{H,\ell}}$. 
Hereafter, in order to ease the notation, we incorporate the bias term in the weight matrix, i.e., $\vec{W}_{\ell}=[\vec{W}_{\ell};\vec{b}_{\ell}]$ and $\vec{h}_{\ell}=[\vec{h}_{\ell};1]$. $\vec{\Theta}=[\vec{W}_{1}, \dots, \vec{W}_{L}]$ groups  all the network's parameters.

Given a training set of $N$ input-output vectors in the ($\vec{x_i}$, $\vec{y_i}$) form, $i=1,\dots,N$, the learning procedure aims at identifying a suitable configuration of parameters $\vec{\hat{\Theta}}$ that minimizes a loss function $\loss$ evaluating the discrepancy between the estimated values $f(\vec{x_t};\vec{\Theta)}$ and the measurements $\vec{y_t}$:

$$
\vec{\hat{\Theta}} = \argmin_\vec{\Theta}. \loss(\vec{\Theta})
$$

The mean squared error:
\begin{equation}
\loss(\vec{\Theta}) = \frac{1}{N} \sum\limits_{t=1}^{N} (\vec{y_t} - f(\vec{x_t};\vec{\Theta)})^2
\label{eq:loss}
\end{equation}
 is a very popular loss function for time series prediction and, not rarely, a regularization penalty term is introduced to prevent overfitting and improve the generalization capabilities of the model
\begin{equation}
\loss(\vec{\Theta}) = \frac{1}{N} \sum\limits_{t=1}^{N} (\vec{y_t} - f(\vec{x_t};\vec{\Theta)})^2 + \mathrm{\Omega}(\vec{\Theta}).
\label{eq:loss_reg}
\end{equation}
The most used regularization scheme controlling model complexity is the L2 regularization $\mathrm{\Omega}(\vec{\Theta}) = \lambda\Vert \vec{\Theta} \Vert_2^2$, being $\lambda$ a suitable hyper-parameter controlling the regularization strength.

As Equation \ref{eq:loss_reg} is not convex, the solution cannot be obtained in a closed form with linear equation solvers or convex optimization techniques. Parameters estimation (learning procedure) operates iteratively e.g., by leveraging on the gradient descent approach:
\begin{equation}
    \vec{\Theta}_k = \vec{\Theta}_{k-1} - \eta \nabla_{\vec{\Theta}}\loss(\vec{\Theta}) \Big\vert_{\vec{\Theta} = \vec{\Theta}_{k-1}}
\end{equation}
where $\eta$ is the learning rate and $\nabla_{\vec{\Theta}}\loss(\vec{\Theta})$ the gradient w.r.t. $\vec{\Theta}$.
Stochastic Gradient Descent (SGD),  RMSProp \cite{RMSProp}, Adagrad \cite{Adagrad}, Adam \cite{Adam} are popular learning procedures.
The learning procedure yields estimate $\vec{\hat{\Theta}} =\vec{\Theta}_k$ associated with the predictive model $f(\vec{x_t};\vec{\hat{\Theta}})$. 

In our work, deep FNNs are the baseline model architectures. 

In multi-step ahead prediction the output layer dimension coincides with the forecasting horizon $\outlen > 1$. The dimension of the input vector depends also on the presence of exogenous variables; this aspect is further discussed in Section \ref{sec:experiments}.

\subsection{Related Work}
The use of Feed Forward Neural networks in short term load forecasting dates back to the 90s.
Authors in \cite{ParkANN} propose a shallow neural network with a single hidden layer to provide a 24-hour forecast using both load and temperature information.  
In \cite{LeeANN} one day ahead forecast is implemented using two different prediction strategies: one network provides all 24 forecast values in a single shot (MIMO strategy) while another  single output network provides the day-ahead prediction by recursively feedbacking its last value estimate (recurrent strategy). The recurrent strategy shows to be more efficient in terms of both training time and forecasting accuracy.
In \cite{ChenFNN} the authors present a feed forward neural network to forecast electric loads on a weekly basis. The sparsely connected feed forward architecture receives the
load time-series, temperature readings, as well as the time and
day of the week. It is shown that the extra information improves the forecast accuracy compared to an ARIMA model trained on the same task.
\cite{SrinANN} presents one of the first multi-layer FNN to forecast the hourly load of a power system.

A detailed review concerning applications of artificial neural networks in short-term load forecasting can be found in \cite{ANNReview}. However, this survey dates back to the early 2000s, and does not discuss deep models. 
More recently, architectural variants of feed forward neural networks have been used; for example, in \cite{ResNet_STLM} a ResNet \cite{ResNet} inspired model is used to provide day ahead forecast by leveraging on a very deep architecture. The article shows a significant improvement on aggregated load forecasting when compared to other (not-neural) regression models on different datasets.

\section{Recurrent Neural networks}
\label{sec:rnn}
In this section we overview recurrent neural networks, and, in particular the Elmann Net architecture \cite{Elman}, Long-Short Term Memory \cite{LSTM} and Gated Recurrent Unit \cite{GRU} networks. Afterwords, we introduce deep recurrent neural networks and discuss different strategies to perform multi-step ahead forecasting.
Finally, we present related work in short-term load forecasting that leverages on recurrent networks.

\subsection{Elmann RNNs (ERNN)}
Elmann Recurrent Neural Networks (ERNN) were proposed in \cite{Elman} to generalize feedforward neural networks for better handling ordered data sequences like time-series. 

The reason behind the effectiveness of RNNs in dealing with sequences of data comes from their ability to learn a compact representation of the input sequence $\vec{x_t}$ by means of a recurrent function $f$ that implements the following mapping:
\begin{equation}
\vec{h}\t{t} = f(\vec{h}\t{t-1}, \vec{x}\t{t}; \vec{\Theta})
\label{eq:rnn_state}
\end{equation}

\begin{figure*}[ht]
    \centering
    \includegraphics[width=.8\linewidth]{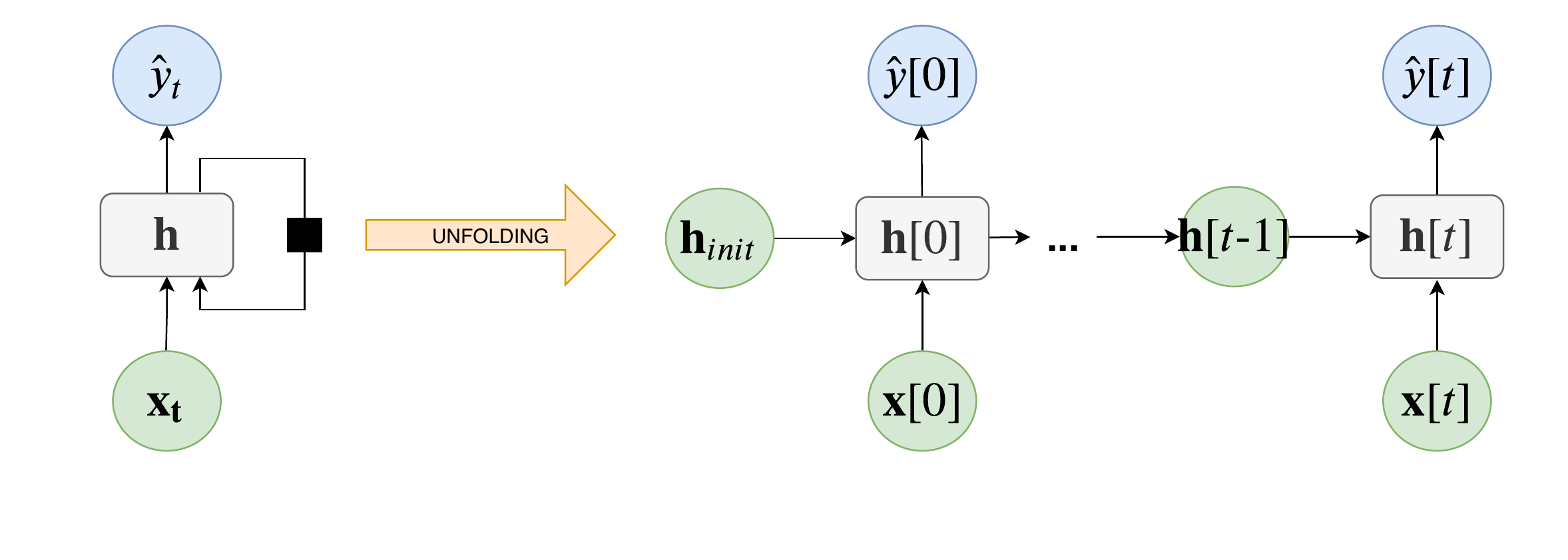}
    \caption{(Left) A simple RNN with a single input. The black box represents the delay operator which leads to Equation \ref{eq:rnn_state}. 
    (Right) The network after unfolding. Note that the structure reminds that of a (deep) feed forward neural network but, here, each layer is constrained to share the same weights. $\vec{h}_{init}$ is the initial state of the network which is usually set to zero.}
\label{fig:RNN_graph}
\end{figure*}

\begin{figure}[H]
\centering
\includegraphics[width=.8\linewidth]{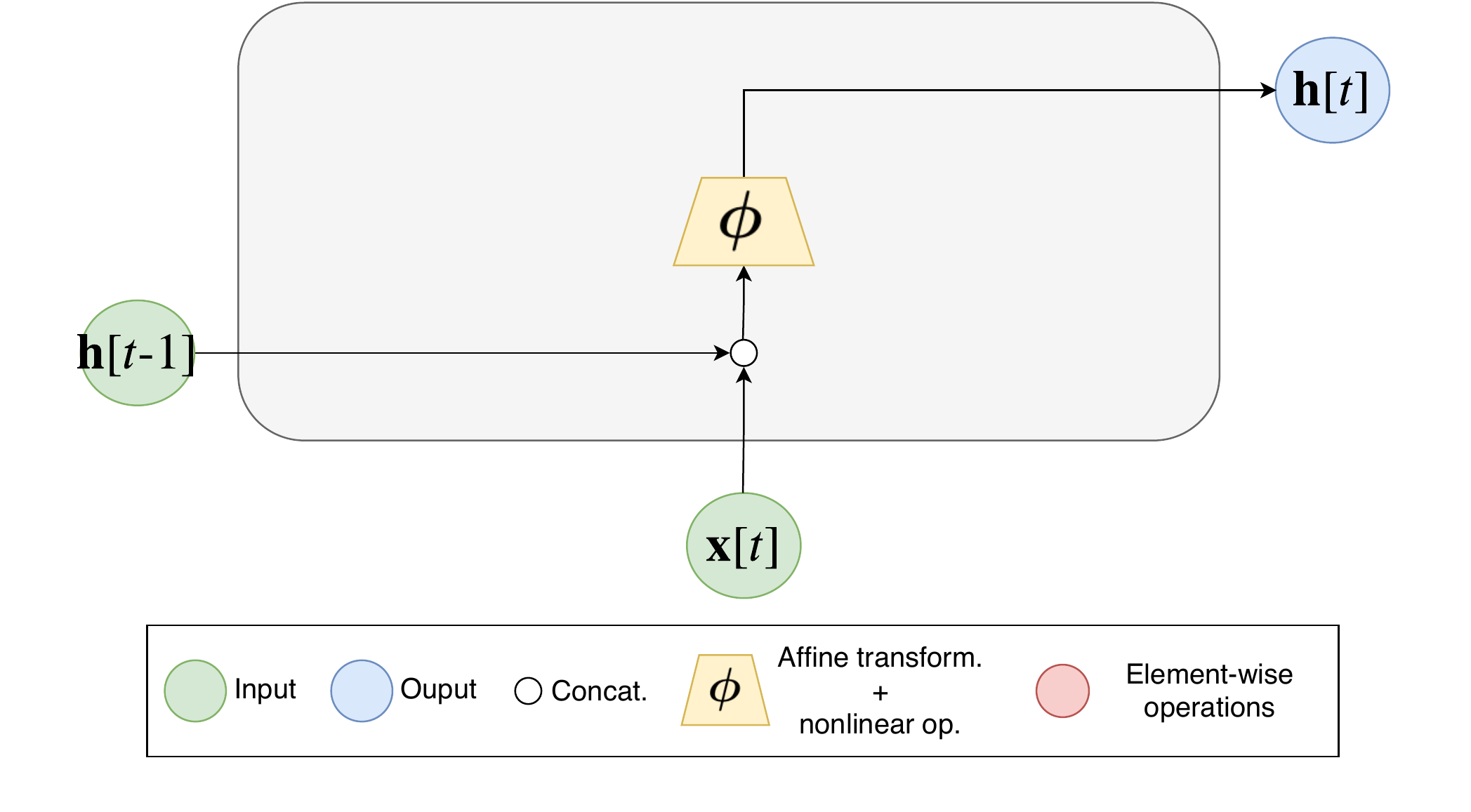}
\caption{A simple ERNN block with one cell implementing Equation \ref{eq:rnn_activation}\ref{eq:rnn_state_up} once rewritten as matrix concatenation: 
$ \vec{a}\t{t} = [\vec{W}, \vec{U}]^T[\vec{h}\t{t-1},\vec{x}\t{t}]$,
$ \vec{h}\t{t} = \phi(\vec{a}\t{t}$),
with $[\vec{W}, \vec{U}] \in \real^{(n_H + d)\times n_H}$ and $[\vec{h}\t{t-1},\vec{x}\t{t}] \in \real^{n_H + d}$,
Usually $\phi(\cdot)$ is the hyperbolic tangent.}
\label{fig:RNN_cell}
\end{figure}

By expanding Equation \ref{eq:rnn_state} and given a sequence of inputs $\vec{x_t} = [\vec{x}\t{0}, \dots , \vec{x}\t{\inlen -1}]$, $\vec{x}\t{t} \in \real^{d}$ the computation becomes:
\begin{align}
\vec{a}\t{t} &= \vec{W}^T\vec{h}\t{t-1} + \vec{U}^T\vec{x}\t{t} \label{eq:rnn_activation}\\
\vec{h}\t{t} &= \phi(\vec{a}\t{t}) \label{eq:rnn_state_up}\\
\vec{y}\t{t} &= \psi(\vec{V}^T\vec{h}\t{t} \label{eq:rnn_out})
\end{align}
where $\vec{W} \in \real^{n_H\times n_H}$, $\vec{U} \in \real^{d \times n_H}$, $\vec{V} \in \real^{n_H \times n_O}$ are the weight matrices for hidden-hidden, input-hidden, hidden-output connections respectively, $\phi(\cdot)$ is an activation function (generally the hyperbolic tangent one) and $\psi(\cdot)$ is normally a  linear function. The computation of a single module in an Elmann recurrent neural network is depicted in Figure \ref{fig:RNN_cell}.

It can be noted that an ERNN processes one element of the sequence at a time, preserving its inherent temporal order.
After reading an element from the input sequence  $\vec{x}\t{t} \in \real^{d}$ the network updates its internal state $\vec{h}\t{t} \in \real^{n_H}$ using both (a transformation of) the latest state $\vec{h}\t{t-1}$ and (a transformation of) the current input (Equation \ref{eq:rnn_state}).
The described process can be better visualized as an acyclic graph obtained from the original cyclic graph (left side of Figure \ref{fig:RNN_graph}) via an operation known as time unfolding (right side of Figure \ref{fig:RNN_graph}). 
It is of fundamental importance to point out that all nodes in the unfolded network share the same parameters, as they are just replicas distributed over time.

The parameters of the network $\vec{\Theta} = [\vec{W},\vec{U}, \vec{V}]$ are usually learned via Backpropagation Through Time (BPTT) \cite{BPTT,BPTT2}, a generalized version of standard Backpropagation. In order to apply gradient-based optimization, the recurrent neural network has to be transformed through the unfolding procedure shown in Figure \ref{fig:RNN_graph}. In this way, the network is converted into a FNN having as many layers as time intervals in the input sequence, and each layer is constrained to have the same weight matrices.
In practice Truncated Backpropagation Through Time \cite{TBPTT_1}  TBPTT($\tau_b$, $\tau_f$) is used. The method processes an input window of length $\inlen$ one timestep at a time and runs BPTT for $\tau_b$ timesteps every $\tau_f$ steps. 
Notice that having $\tau_b < \inlen$ does not limit the memory capacity of the network as the hidden state incorporates information taken from the whole sequence. Despite that, setting $\tau_b$ to a very low number may result in poor performance.
In the literature BPTT is considered equivalent to TBPTT($\tau_b=\inlen$, $\tau_f=1$).
In this work we used epoch-wise Truncated BPTT i.e., TBPTT($\tau_b=\inlen$, $\tau_f=\inlen$) to indicate that the weights update is performed once a whole sequence has been processed.

Despite of the model simplicity, Elmann RNNs are hard to train due to ineffectiveness of gradient (back)propagation.  In fact, it emerges that the propagation of gradient is effective for short-term connections but is very likely to fail for long-term ones, when the gradient norm usually shrinks to zero or diverges. These two behaviours are known as the vanishing gradient and the exploding gradient problems \cite{VanishExpGrad, TrainRNN} and were extensively studied in the machine learning community. 

\subsection{Long Short-Term Memory (LSTM)}
Recurrent neural networks with Long Short-Term Memory (LSTM) were introduced to cope with the vanishing and exploding gradients problems occurring in ERNNs and, more in general, in standard RNNs \cite{LSTM}.
LSTM networks maintain the same topological structure of ERNN but differ in the composition of the inner module - or cell.

\begin{figure}[H]
\centering
\includegraphics[width=.8\linewidth]{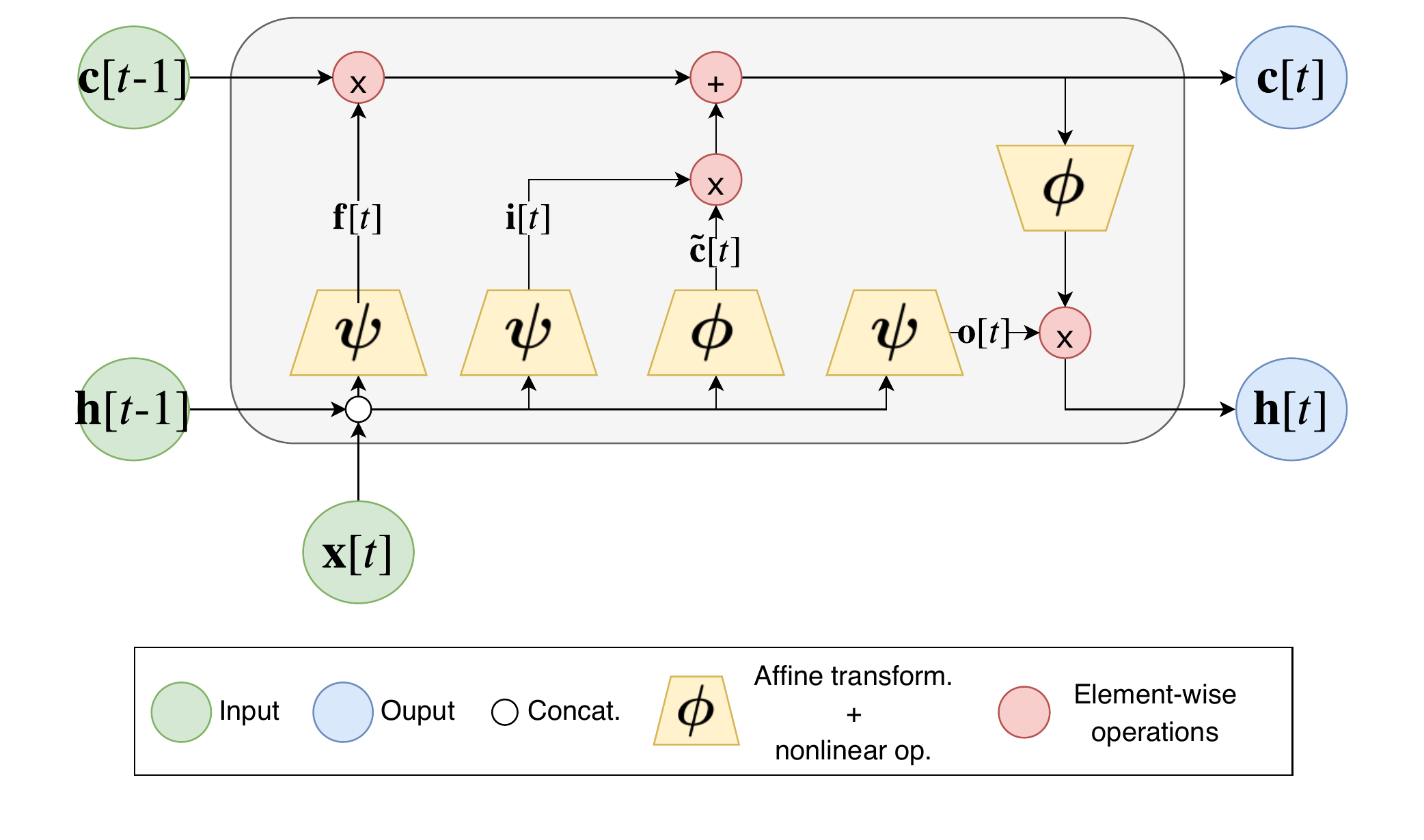}
\caption{Long-Short Term Memory block with one cell.}
\label{fig:LSTM_cell}
\end{figure}

Each LSTM cell has the same input and output as an ordinary ERNN cell but, internally, it implements a gated system that controls the neural information processing (see Figure Figure \ref{fig:RNN_cell} and \ref{fig:LSTM_cell}).
The key feature of gated networks is their ability to control the gradient flow by acting on the gate values; this allows to tackle the vanishing gradient problem, as LSTM can maintain its internal memory unaltered for long time intervals. Notice from the equations below that the inner state of the network results as a linear combination of the old state and the new state (Equation \ref{eq:gated_lc}). Part of the old state is preserved and flows forward while in the ERNN the state value is completely replaced at each timestep (Equation \ref{eq:rnn_state_up}). In detail, the neural computation is:
\begin{align}
\vec{i\t{t}} &= \psi\left( \vec{W_f}\vec{h}\t{t-1} +  \vec{U_f}\vec{x}\t{t} \right) \\
\vec{f\t{t}} &= \psi\left( \vec{W_i}\vec{h}\t{t-1} +  \vec{U_i}\vec{x}\t{t} \right) \\
\vec{o\t{t}} &= \psi\left( \vec{W_o}\vec{h}\t{t-1} +  \vec{U_o}\vec{x}\t{t} \right) \\
\vec{\widetilde{c}\t{t}} &= \phi\left( \vec{W_c}\vec{h}\t{t-1} +  \vec{U_c}\vec{x}\t{t} \right)\\
\vec{c\t{t}} &= \vec{f}\t{t}\odot\vec{c}\t{t-1} + \vec{i}\t{t}\odot\vec{\widetilde{c}}\t{t}  \label{eq:gated_lc}\\ 
\vec{h\t{t}} &= \vec{o}\t{t} \odot \phi(\vec{c}\t{t})
\end{align}
where $\vec{W_f}, \vec{W_i}, \vec{W_o}, \vec{W_c} \in \real^{n_H \times n_H}$,
$\vec{U_f}, \vec{U_i}, \vec{U_o}, \vec{U_c} \in \real^{n_H \times d}$ are parameters to be learned,
$\odot$ is the Hadamard product, 
$\psi(\cdot)$ is generally a sigmoid activation while $\phi(\cdot)$ can be any non-linear one (hyperbolic tangent in the original paper).
The cell state $\vec{c}\t{t}$ encodes the - so far learned -  information from the input sequence. At timestep $t$ the flow of information within the unit is controlled by three elements called gates: the forget gate $\vec{f}\t{t}$ controls the cell state's content and changes it when obsolete, the input gate $\vec{i}\t{t}$ controls which state value will be updated and how much, $\vec{\widetilde{c}}\t{t}$, finally the output gate $\vec{o}\t{t}$  produces a filtered version of the cell state and serves it as the network's output $\vec{h}\t{t}$ \cite{LSTMSpaceOdyssey}.

\subsection{Gated Recurrent Units (GRU)}
Firstly introduced in \cite{GRU}, GRUs are a simplified variant of LSTM and, as such, belong to the family of gated RNNs. GRUs distinguish themselves from LSTMs for merging in one gate functionalities controlled by the forget gate and the input gate. This kind of cell ends up having just two gates, which results in a more parsimonious architecture compared to LSTM that, instead, has three gates. 

\begin{figure}[H]
\centering
\includegraphics[width=.8\linewidth]{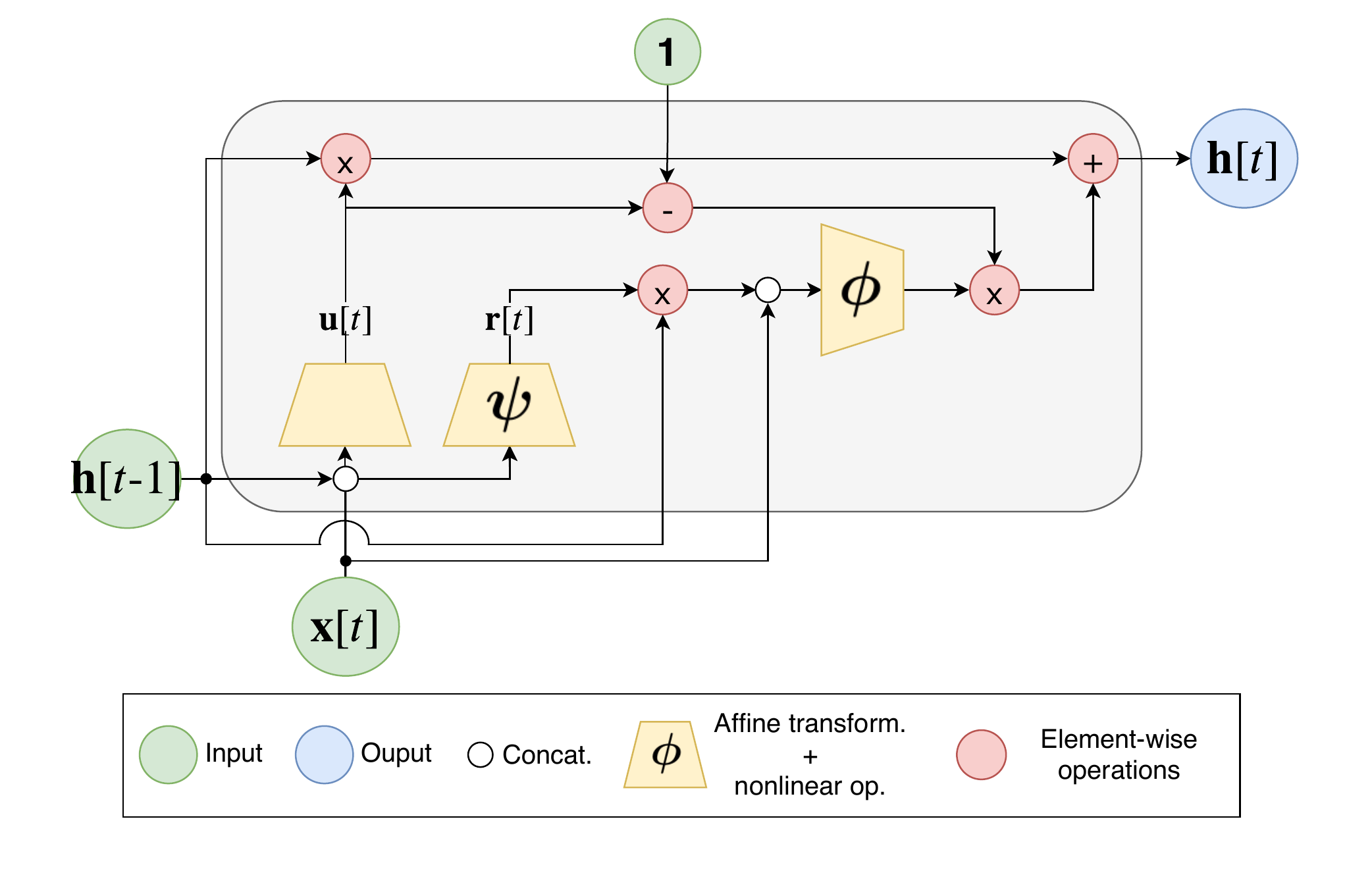}
\caption{Gated Recurrent Unit memory block with one cell.}
\label{fig:GRU_cell}
\end{figure}

The basic components of a GRU cell are outlined in Figure \ref{fig:GRU_cell}, 
whereas the neural computation is controlled by:
\begin{align}
\vec{u\t{t}} &= \psi\left( \vec{W_u}\vec{h}\t{t-1} +  \vec{U_u}\vec{x}\t{t} \right) \\
\vec{r\t{t}} &= \psi\left( \vec{W_r}\vec{h}\t{t-1} +  \vec{U_r}\vec{x}\t{t} \right) \\
\vec{\widetilde{h}\t{t}} &= \phi\left( \vec{W_c}\left(\vec{r}\t{t}\odot\vec{h}\t{t-1} \right) +  \vec{U_c}\vec{x}\t{t} \right)\\
\vec{h\t{t}} &= \vec{u\t{t}}\odot\vec{h}\t{t-1} + \left( 1 - \vec{u}\t{t} \right)\odot \vec{\widetilde{h}}\t{t}
\end{align}
where $\vec{W_u}, \vec{W_r}, \vec{W_c} \in \real^{n_H \times n_H}$,
$\vec{U_u}, \vec{U_r}, \vec{U_c} \in \real^{n_H \times d}$ are the parameters to be learned,
$\psi(\cdot)$ is generally a sigmoid activation while $\phi(\cdot)$ can be any kind of non-linearity (in the original work it was an hyperbolic tangent).
$\vec{u}\t{t}$ and $\vec{r}\t{t}$ are the update and the  reset gates, respectively. 
Several works in the natural language processing community show that GRUs perform comparably to LSTM but train generally faster due to the lighter computation \cite{GRU_Eval, GRUvsLSTM}.

\subsection{Deep Recurrent Neural Networks}
\label{sec:deeprnn}
All recurrent architectures presented so far are characterized by a single layer. In turn, this implies that the computation is composed by an affine transformation followed by a non-linearity.
That said, the concept of depth in RNN is less straightforward than in feed-forward architectures. Indeed, the later ones become deep when the input is processed by a large number of non-linear transformations before generating the output values. 
However, according to this definition, an unfolded RNN is already a deep model given its multiple non-linear processing layers.
That said, a deep multi-level processing can be applied to all the transition functions (input-hidden, hidden-hidden, hidden-output) as there are no intermediate layers involved in these computations \cite{Pascanu}.
Deepness can also be introduced in recurrent neural networks by stacking recurrent layers one on top of the other \cite{StackedRNN}. As this deep architecture is more intriguing, in this work, we refer it as a Deep RNN. By iterating the RNN computation, the  function implemented by the deep architecture can be represented as:
\begin{equation}
\vec{h}_{\ell}\t{t} = f(\vec{h}_{\ell}\t{t-1}, \vec{h}_{\ell-1}\t{t}; \vec{\Theta}) \qquad \ell = {1,2,...,L}
\end{equation}
where $\vec{h}_{\ell}\t{t}$ is the hidden state at timestep $t$ for layer $\ell$. Notice that $\vec{h}_0\t{t} = \vec{x}\t{t}$.
It has been empirically shown in several works that Deep RNNs are better to capture the temporal hierarchy exhibited by time-series then their shallow counterpart \cite{Pascanu,Graves,DeepRNN}. Of course, hybrid architectures having different layers -recurrent or not- can be considered as well.

\subsection{Multi-Step Prediction Schemes}
\label{sec:rnn-ms}
There are five different architecture-independent strategies for multi-step ahead forecasting \cite{ReviewMultiStep}:

\paragraph*{Recursive strategy (Rec)} a single model is trained to perform a one-step ahead forecast given the input sequence.
Subsequently, during the operational phase, the forecasted output is recursively fedback and considered to be the correct one. By iterating $\outlen$ times this procedure we generate the forecast values at time $t+\outlen$.
The procedure is described in Algorithm \ref{alg:rec}, where $\vec{x}[1:]$ is the input vector without its first element while the $vectorize(\cdot)$ procedure concatenates the scalar output $y$ to the exogenous input variables.
\begin{algorithm}
\caption{Recursive Strategy (Rec) for Multi-Step Forecasting}
\begin{algorithmic}[1]
\STATE{$\vec{x} \gets \vec{x_t}$}
\STATE{$\vec{o} \gets $ empty list}
\STATE{$k \gets 1$}
\WHILE{$k < \outlen+1$}
    \STATE{$o \gets f(\vec{x})$}
    \STATE{$\vec{o} \gets concatenate(\vec{o}, o)$}
    \STATE{$\vec{x} \gets concatenate(\vec{x}[1:], vectorize(o)))$}
    \STATE{$k \gets k+1$}
\ENDWHILE
\RETURN{$\vec{o}$ as $\vec{\hat{y}_t}$}
\end{algorithmic}
\label{alg:rec}
\end{algorithm}

To summarize, the predictor $f$ receives in input a vector $\vec{x}$ of length $\inlen$ and outputs a scalar value $o$.

\paragraph*{Direct strategy} design a set of $\outlen$ independent predictors $f_k, k=1, \dots, \outlen$, each of which providing a forecast at time $t + k$. Similarly to the recursive strategy, each predictor $f_k$ outputs a scalar value $o$, but the input vector is the same to all the predictors.
Algorithm \ref{alg:dir} details the procedure.
\begin{algorithm}
\caption{Direct Strategy for Multi-Step Forecasting}
\begin{algorithmic}[1]
\STATE{$\vec{x} \gets \vec{x_t}$}
\STATE{$\vec{o} \gets $ empty list}
\STATE{$k \gets 1$}
\WHILE{$k < \outlen+1$}
    \STATE{$\vec{o} \gets concatenate(\vec{o}, f_k(\vec{x}))$}
    \STATE{$k \gets k+1$}
\ENDWHILE
\RETURN{$\vec{o}$ as $\vec{\hat{y}_t}$}
\end{algorithmic}
\label{alg:dir}
\end{algorithm}

\paragraph*{DirRec strategy} \cite{DirRec} is a combination of the above two strategies. Similar to the direct approach, $\outlen$ models are used, but here, each predictor leverages on an enlarged input set, obtained by adding the results of the forecast at the previous timestep. The procedure is detailed in Algorithm \ref{alg:direc}.
\begin{algorithm}
\caption{DirRec Strategy for Multi-Step Forecasting}
\begin{algorithmic}[1]
\STATE{$\vec{x} \gets \vec{x_t}$}
\STATE{$\vec{o} \gets $ empty list}
\STATE{$k \gets 1$}
\WHILE{$k < \outlen+1$}
    \STATE{$o \gets f_k(\vec{x})$}
    \STATE{$\vec{o} \gets concatenate(\vec{o}, o)$}
    \STATE{$\vec{x} \gets concatenate(\vec{x}, vectorize(o))$}
    \STATE{$k \gets k+1$}
\ENDWHILE
\RETURN{$\vec{o}$ as $\vec{\hat{y}_t}$}
\end{algorithmic}
\label{alg:direc}
\end{algorithm}

\paragraph*{MIMO strategy}(Multiple input - Multiple output) \cite{MIMO}, a single predictor $f$ is trained to forecast a whole output sequence of length $n_O$ in one-shot, i.e., differently from the previous cases the output of the model is not a scalar but a vector:
\begin{align*}
\vec{\hat{y}_t} = f(\vec{x_t})
\end{align*}

\paragraph*{DIRMO strategy} \cite{DIRMO}, represents a trade-off between the Direct strategy and the MIMO strategy. It divides the $\outlen$ steps forecasts into smaller forecasting problems, each of which of length $s$. It follows that $\lceil \frac{\outlen}{s} \rceil$ predictors are used to solve the problem.

Given the considerable computational demand required by RNNs during training, we focus on multi-step forecasting strategies that are computationally cheaper, specifically, Recursive and MIMO strategies \cite{ReviewMultiStep}. We will call them RNN-Rec and RNN-MIMO.

Given the hidden state $\vec{h}\t{t}$ at timestep $t$, the hidden-output mapping is obtained through a fully connected layer on top of the recurrent neural network. The objective of this dense network is to learn the mapping between the last state of the recurrent network, which represents a kind of lossy summary of the task-relevant aspect of the input sequence, and the output domain.
This holds for all the presented recurrent networks and is consistent with Equation \ref{eq:rnn_out}. 
In this work RNN-Rec and RNN-MIMO differ in the cardinality of the output domain, which is $1$ for the former and $\outlen$ for the latter, meaning that in Equation \ref{eq:rnn_out} either $\vec{V} \in \real^{n_H \times 1}$ or $\vec{V} \in \real^{n_H \times \outlen}$. 
The objective function is:
\begin{equation}
    \loss(\vec{\Theta}) = \frac{1}{\outlen}\sum\limits_{t=0}^{\outlen-1} (y\t{t} - \hat{y}\t{t})^2 + \mathrm{\Omega}(\vec{\Theta})
\end{equation}

\subsection{Related work}
\label{sec:rnn_art}
In \cite{ElmannSTLF} an Elmann recurrent neural network is considered to provide hourly load forecasts. The study also compares the performance of the network when additional weather information such as temperature and humidity are fed to the model. The authors conclude that, as expected, the recurrent network benefits from multi-input data and, in particular, weather ones. \cite{ElmannPalermo} makes use of ERNN to forecast household electric consumption obtained from a suburban area in the neighbours of Palermo (Italy). In addition to the historical load measurements, the authors introduce several features to enhance the model's predictive capabilities. Besides the weather and the calendar information, a specific ad-hoc index was created to assess the influence of the use of air-conditioning equipment on the electricity demand.
In recent years, LSTMs have been adopted in short term load forecasting, proving to be more effective then traditional time-series analysis methods. In \cite{ESTLF_LSTM} LSTM is shown to outperform traditional forecasting methods being able to exploit the long term dependencies in the time series to forecast the day-ahead load consumption.
Several works proved to be successful in enhancing the recurrent neural network capabilities by employing multivariate input data. In \cite{STLF_LSTM} the authors propose a deep, LSTM based architecture that uses past measurements of the whole household consumption along with some measurements from selected appliances to forecast the consumption of the subsequent time interval (i.e., a one step prediction). In \cite{GA_LSTM} a LSTM-based network is trained using a multivariate input which includes temperature, holiday/working day information, date and time information. 
Similarly, in \cite{PowerLSTM} a power demand forecasting model based on LSTM shows an accuracy improvement compared to more traditional machine learning techniques such as Gradient Boosting Trees and Support Vector Regression.

GRUs have not been used much in the literature as LSTM networks are often preferred. That said, the use of GRU-based networks is reported in \cite{RNN_Bianchi}, while a more recent study \cite{GRU_STLF} uses GRUs for the daily consumption forecast of individual customers.
Thus, investigating deep GRU-based architectures is a relevant scientific topic, also thanks to their faster convergence and simpler structure compared to LSTM \cite{GRU_Eval}.

Despite all these promising results, an extensive study of recurrent neural networks \cite{RNN_Bianchi}, and in particular of ERNN, LSTM, GRU, ESN\cite{ESN} and NARX, concludes that none of the investigated recurrent architectures manages to outperform the others in all considered experiments. Moreover, the authors noticed that recurrent cells with gated mechanisms like LSTM and GRU perform comparably well to much simpler ERNN. This may indicate that in short-term load forecasting gating mechanism may be unnecessary; this issue is further investigated -and evidence found- in the present work.

\section{Sequence To Sequence models}
Sequence To Sequence (seq2seq) architectures \cite{Seq2Seq} or encoder-decoder models \cite{GRU} were initially designed to solve RNNs inability to produce output sequences of arbitrary length. The architecture was firstly used in neural machine translation \cite{Seq2Seq_NMT,Seq2Seq_NMT1, GRU} but has emerged as the golden standard in different fields such as speech recognition \cite{Seq2Seq_SpeechRec,Seq2Seq_SpeechRec_Att1, Seq2Seq_SpeechRec_Att} and image captioning \cite{Seq2Seq_ImgCaptioning}.

\begin{figure*}[t]
    \centering
    \includegraphics[width=\textwidth]{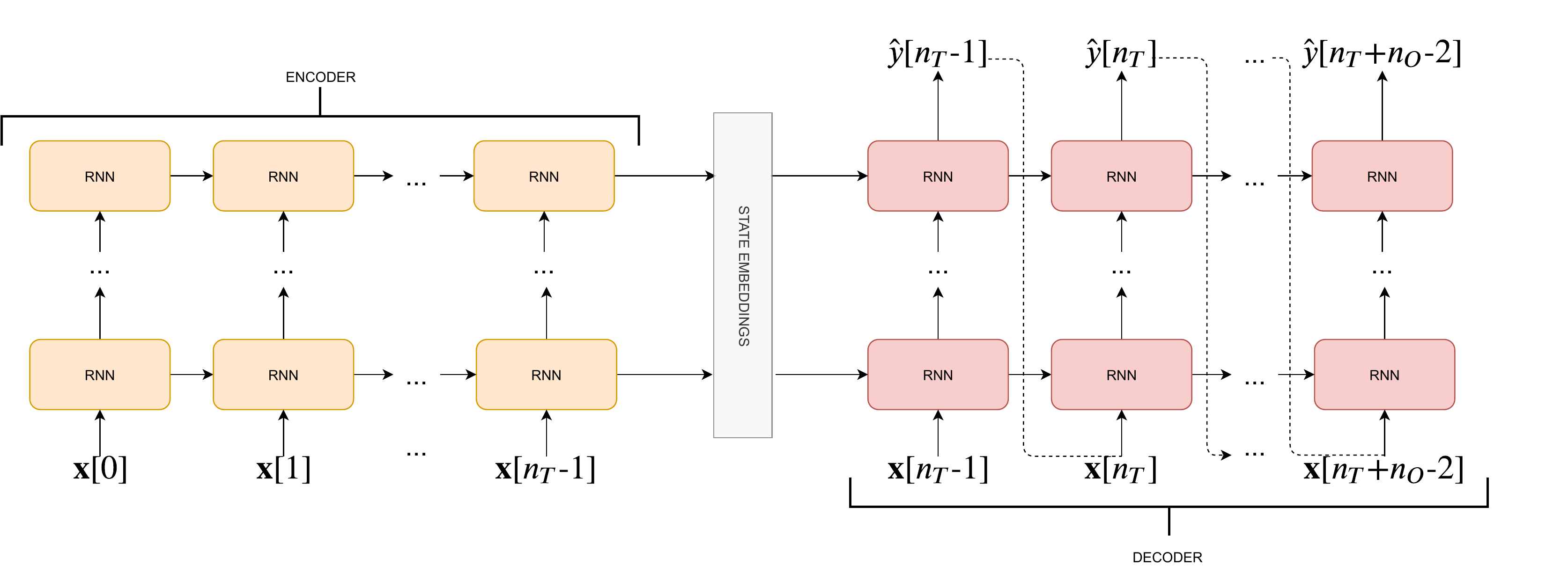}
    \caption{seq2seq (Encoder-Decoder) architecture with a general Recurrent Neural network both for the encoder and the decoder. Assuming a Teacher Forcing training process, the solid lines in the decoder represents the training phase while the dotted lines depicts the values'path during prediction.}
    \label{fig:S2S}
\end{figure*}

The core idea of this general framework is to employ two networks resulting in an encoder-decoder architecture.
The first neural network (possibly deep) $f$, an encoder, reads the input sequence $\vec{x_t} \in \real^{\inlen \times d}$ of length $\inlen$ one timestep at a time; the computation generates a, generally lossy, fixed dimensional vector representation of it $\vec{c} = f(\vec{x_t}, \vec{\Theta}_f)$, $\vec{c} \in \real^{d'}$. This embedded representation is usually called context in the literature and can be the last hidden state of the encoder or a function of it.
Then, a second neural network $g$ - the decoder - will learn how to produce the output sequence $\vec{\hat{y}_t} \in \real^{\outlen}$ given the context vector, i.e., $\vec{\hat{y}}=g(\vec{c}, \vec{\Theta}_g)$. The schematics of the whole architecture is depicted in Figure \ref{fig:S2S}.

\label{sec:s2s}

The encoder and the decoder modules are generally two recurrent neural networks trained end-to-end to minimize the objective function:
\begin{equation}
\loss(\vec{\Theta}) = \sum\limits_{t=0}^{\outlen-1} (y\t{t} - \hat{y}\t{t})^2  + \mathrm{\Omega}(\vec{\Theta}), \quad \vec{\Theta} = [\vec{\Theta}_f, \vec{\Theta}_g]
\end{equation}
\begin{equation}
\hat{y}\t{t} = g(y\t{t-1}, \vec{h}\t{t-1}, \vec{c} ; \vec{\Theta})
\label{eq:s2s_train_tf}
\end{equation}
where $\hat{y}\t{t}$ is the decoder's estimate at time $t$, $y\t{t}$ is the real measurement, $\vec{h}\t{t-1}$ is the decoder's last state, $\vec{c}$ is the context vector from the encoder, $\vec{x}$ is the input sequence and $\mathrm{\Omega}(\vec{\Theta})$ the regularization term.
The training procedure for this type of architecture is called teacher forcing \cite{TeacherForcing}.
As shown in Figure \ref{fig:S2S} and explained in Equation \ref{eq:s2s_train_tf}, during training, the decoder's input at time $t$ is the ground-truth value $y\t{t-1}$, which is then used to generate the next state $\vec{h}\t{t}$ and, then, the estimate $\hat{y}\t{t}$. During inference the true values are unavailable and replaced by the estimates: 
\begin{equation}
\hat{y}\t{t} = g(\hat{y}\t{t-1}, \vec{h}\t{t-1}, \vec{c}; \vec{\Theta}).
\label{eq:s2s_train_sg}
\end{equation}
This discrepancy between training and testing results in errors accumulating over time during inference. 
In the literature this problem is often referred to as exposure bias \cite{ExposureBias}.
Several solutions have been proposed to address this problem; in \cite{scheduledSampling} the authors present scheduled sampling, a curriculum learning strategy that gradually changes the training process by switching the decoder's inputs from ground-truth values to model's predictions. 
The \textit{professor forcing} algorithm, introduced in \cite{ProfessorForcing}, uses an adversarial framework to encourage the dynamics of the recurrent network to be the same both at training and operational (test) time.
Finally, in recent years, reinforcement learning methods have been adopted to train sequence to sequence models; a comprehensive review is presented in \cite{RL_S2S}.

\begin{figure*}[t]
    \centering
    \includegraphics[width=0.9\textwidth]{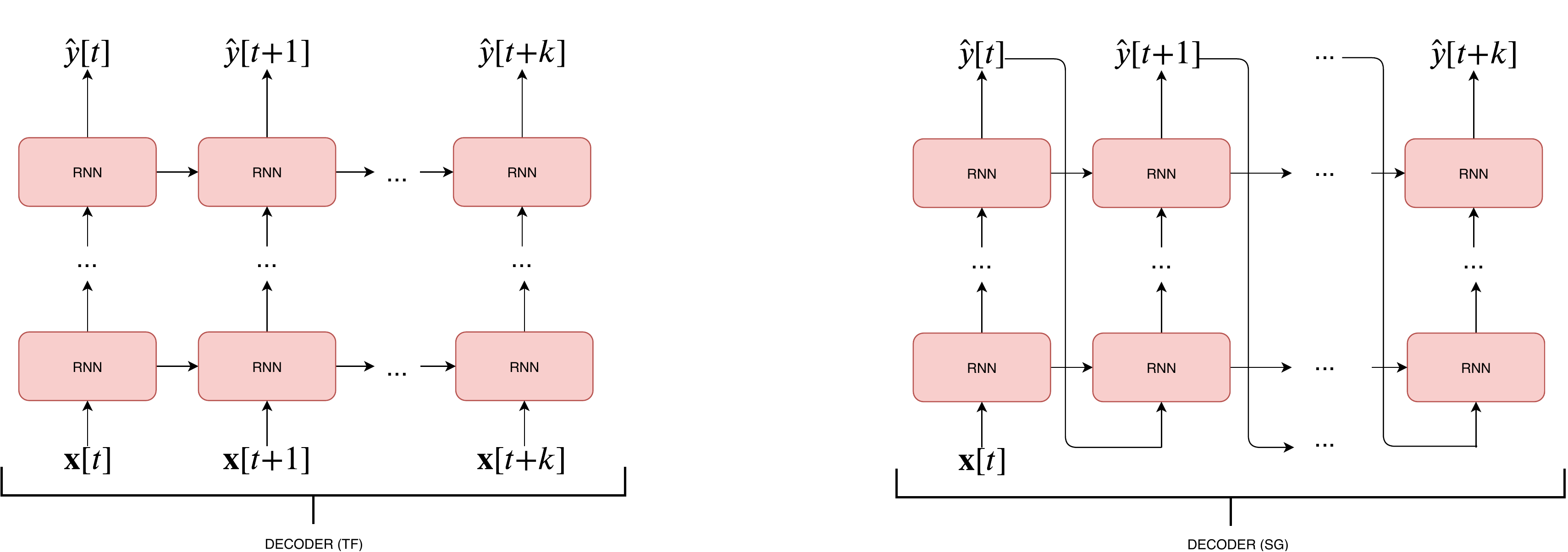}
    \caption{(Left) decoder with ground-truth inputs (Teacher Forcing). (Right) Decoder with self-generated inputs.}
    \label{fig:seq2seq_tf_sg}
\end{figure*}

In this work we investigate two sequence to sequence architectures, one trained via \textit{teacher forcing} (TF) and one using \textit{self-generated} (SG) samples. The former is characterized by Equation \ref{eq:s2s_train_tf} during training while Equation \ref{eq:s2s_train_sg} is used during prediction. The latter architecture adopts Equation \ref{eq:s2s_train_sg} both for training and prediction. The decoder's dynamics are summarized in Figure \ref{fig:seq2seq_tf_sg}.  It is clear that the two training procedures differ in the  decoder's input source: ground-truth values in teacher forcing, estimated values in self-generated training.

\subsection{Related Work}
Only recently seq2seq models have been adopted in short term load forecasting. In \cite{DLMarino} a LSTM based encoder-decoder model is shown to produce superior performance compared to standard LSTM. In \cite{Seq2SeqGEFCom2014} the authors introduce an adaptation of RNN based sequence-to-sequence architectures for time-series forecasting of electrical loads to demonstrate its better performance with respect to a suite of models ranging from standard RNNs to classical time series techniques.

\section{Convolutional Neural Networks}
\label{sec:cnn}
Convolutional Neural Networks (CNNs) \cite{LeNet} are a family of neural networks designed to work with data that can be structured in a grid-like topology. CNNs were originally used on two dimensional and three-dimensional images, but they are also suitable for one-dimensional data such as univariate time-series.
Once recognized as a very efficient solution for image recognition and classification \cite{AlexNet, VGG,Inception,ResNet}, CNNs have experienced wide adoption in many different computer vision tasks \cite{FastRCNN, FasterRCNN, STN, PRSR, SRGAN}.
Moreover, sequence modeling tasks, like short term electric load forecasting, have been mainly addressed with recurrent neural networks, but recent research indicates that convolutional networks can also attain state-of-the-art-performance in several applications including audio generation \cite{WaveNet}, machine translation \cite{ConvMT} and time-series prediction \cite{ConditionalCNN}.

As the name suggests, these kind of networks are based on a discrete convolution operator that produces an output feature map $\vec{f}$ by sliding a kernel $\vec{w}$ over the input $\vec{x}$.
Each element in the output feature map is obtained by summing up the result of the element-wise multiplication between the input patch (i.e., a slice of the input having the same dimensionality of the kernel) and the kernel. The number of kernels (filters) $M$ used in a convolutional layer determines the depth of the output volume (i.e., the number of output feature maps). To control the other spatial dimensions of the output feature maps two hyper-parameters are used: stride and padding. Stride represents the distance between two consecutive input patches and can be defined for each direction of motion. Padding refers to the possibility of implicitly enlarging the inputs by adding (usually) zeros at the borders to control the output size w.r.t the input one. Indeed, without padding, the dimensionality of the output would be reduced after each convolutional layer.


Considering a 1D time-series $\vec{x} \in \real^{\inlen}$ and a one-dimensional kernel $\vec{w} \in \real^{k}$, the $i^{th}$ element of the convolution between $\vec{x}$ and $\vec{w}$ is:

\begin{equation}
f(i) = (\vec{x} * \vec{w})(i) = \sum\limits_{j=0}^{k-1} x(i-j) w(j)
\label{eq:conv1d}
\end{equation}

with $\vec{f} \in \real^{\inlen-k+1}$ if no zero-padding is used, otherwise padding matches the input dimensionality, i.e., $\vec{f} \in \real^{\inlen}$. 
Equation \ref{eq:conv1d} is referred to the one-dimensional input case but can be easily extended to multi-dimensional inputs (e.g., images, where $\vec{x} \in \real^{W \times H \times D}$) \cite{ConvMath}.
The reason behind the success of these networks can be summarized in the following three points:
\begin{itemize}
\item local connectivity: each hidden neuron is connected to a subset of input neurons that are 
close to each other (according to specific spatio-temporal metric). This property allows the network to drastically reduce the number of parameters to learn (w.r.t. a fully connected network) and facilitate computations. 
\item parameter sharing: the weights used to compute the output neurons in a feature map are the same, so that the same kernel is used for each location. This allows to reduce the number of parameters to learn.
\item translation equivariance: the network is robust to an eventual shifting of its input.
\end{itemize}

In our work we focus on a convolutional  architecture inspired by Wavenet \cite{WaveNet}, a fully probabilistic and autoregressive model used for generating raw audio wave-forms and extended to time-series prediction tasks \cite{ConditionalCNN}.
Up to the authors' knowledge this architecture has never been proposed to forecast the electric load.
A recent empirical comparison between temporal convolutional networks and recurrent networks has been carried out in \cite{TCNEval} on tasks such as polymorphic music and charter-sequence level modelling. The authors were the first to use the name Temporal Convolutional Networks (TCNs) to indicate convolutional networks which are autoregressive, able to process sequences of arbitrary length and output a sequence of the same length. To achieve the above the network has to employ causal (dilated) convolutions and residual connections should be used to handle a very long history size.

\begin{figure*}[t]
    \hspace{-0.1cm}
    \begin{minipage}[t]{0.4\linewidth}
            \centering
            \includegraphics[width=0.8\textwidth]{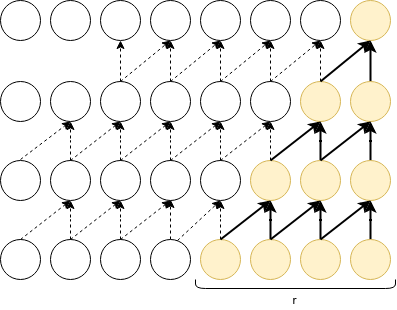}
            \caption{A  3 layers CNN with causal convolution (no dilation), the receptive field $r$ is 4.}
    \end{minipage}
    \hspace{0.3cm}
    \begin{minipage}[t]{0.53\linewidth}
            \includegraphics[width=1.\textwidth]{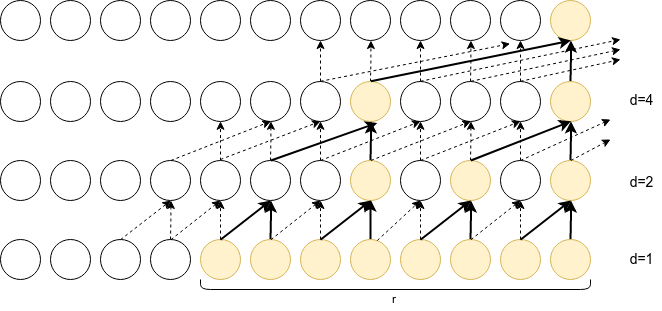}
            \vspace{-0.38cm}
            \caption{A 3 layers CNN with dilated causal convolutions. The dilation factor $d$ grows on each layer by a factor of two and the kernel size $k$ is 2, thus the output neuron is influence by 8 input neurons, i.e., the history size is 8}
        \label{fig:dilated}
    \end{minipage}
\end{figure*}

\paragraph{Dilated Causal Convolution (DCC)} 
Being TCNs a family of autoregressive models, the estimated value at time $t$ must depend only on past samples and not on future ones (Figure \ref{fig:dilated}). To achieve this behavior in a Convolutional Neural Network the standard convolution operator is replaced by causal convolution. 
Moreover, zero-padding of length (filter size - 1) is added to ensure that each layer has the same length of the input layer. 
To further enhance the network capabilities \textit{dilated causal convolutions} are used, allowing to increase the receptive field of the network (i.e., the number of input neurons to which the filter is applied) and its ability to learn long-term dependencies in the time-series. 
Given a one-dimensional input $\vec{x} \in \real^{\inlen}$, and a kernel $\vec{w} \in \real^k$, a dilated convolution output using a dilation factor $d$ becomes:

\begin{equation}
f(i) = (\vec{x} *_d \vec{w})(i) = \sum\limits_{j=0}^{k-1} x(i-dj) w(j)
\end{equation}
This is a major advantage w.r.t simple causal convolutions, as in the later case the receptive field $r$ grows linearly with the depth of the network $r=k(L-1)$ while with dilated convolutions the dependence is exponential $r=2^{L-1}k$, ensuring that a much larger history size is used by the network.

\paragraph{Residual Connections}
Despite the implementation of dilated convolution, the CNN still needs a large number of layers to learn the dynamics of the inputs. 
Moreover, performance often degrade with the increase of the network depth.
The degradation problem has been first addressed in \cite{ResNet} where the authors propose a deep residual learning framework. The authors observe that for a $L$-layers network with a training error $\epsilon$, inserting $k$ extra layers on top of it should either leave the error unchanged or improve it. Indeed, in the worst case scenario, the new $k$ stacked non linear layers should learn the identity mapping $\vec{y} = \mathcal{H}(\vec{x}) = \vec{x}$ where $\vec{x}$ is the output of the network having $L$ layers and $\vec{y}$ is the output of the network with $L + k$ layers. 
Although almost trivial, in practice, neural networks experience problems in learning this identity mapping. The proposed solution suggests these stacked layers to fit a residual mapping $\mathcal{F}(\vec{x}) = \mathcal{H}(\vec{x}) - \vec{x}$ instead of the desired one, $\mathcal{H}(\vec{x})$.
The original mapping is recast into $\mathcal{F}(\vec{x})+\vec{x}$ which is realized by feed forward neural networks with shortcut connections; in this way the identity mapping is learned by simply driving the weights of the stacked layers to zero. 

\begin{figure*}[t]
    \centering
    \includegraphics[width=\linewidth]{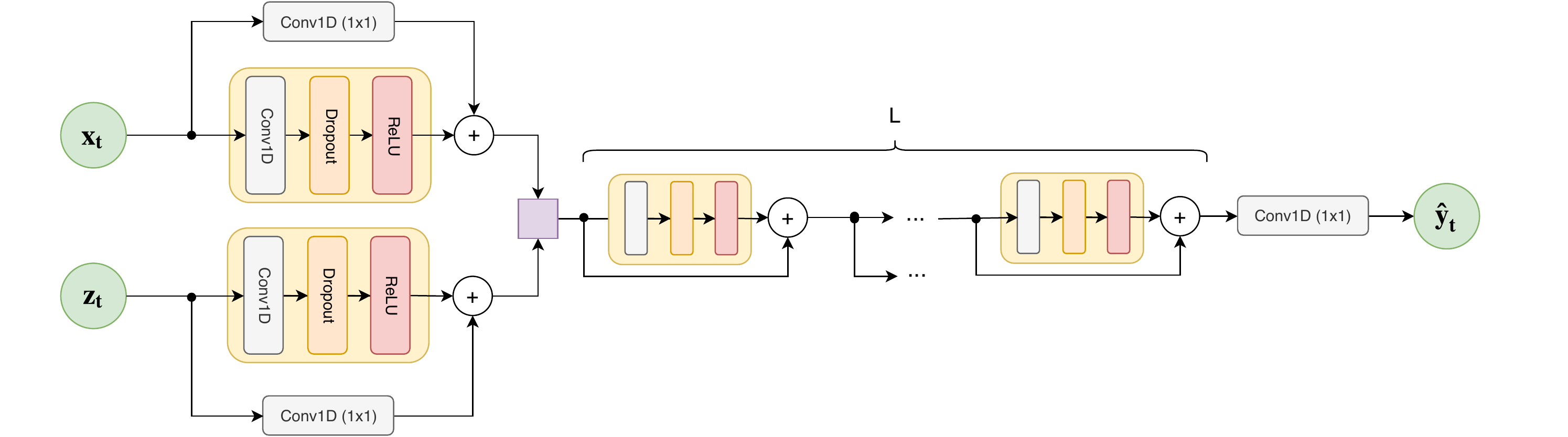}
    \caption{TCN Architecture. $\vec{x_t}$ is the vector of historical loads along with the exogenous features for the time window indexes from $0$ to $\inlen$, $\vec{z_t}$ is the vector of exogenous variables related to the last $\outlen$ indexes of the time window (when available), $\vec{\hat{y}_t}$ is the output vector. Residual Blocks are composed by a 1D Dilated Causal Convolution, a ReLU activation and Dropout. The square box represents a concatenation between (transformed) exogenous  features and (transformed) historical loads.}
    \label{fig:tcn}
\end{figure*}

By means of the two aforementioned principles, the temporal convolutional network is able to exploit a large history size in an efficient manner. Indeed, as observed in \cite{TCNEval}, these models present several computational advantages compared to RNNs. In fact, they have lower memory requirements during training and the predictions for later timesteps are not done sequentially but can be computed in parallel exploiting parameter sharing. Moreover, TCNs training is much more stable than that involving RNNs allowing to avoid the exploding/vanishing gradient problem. 
For all the above, TCNs have demonstrated to be promising area of research for time series prediction problems and here, we aim to assess their forecasting performance w.r.t state-of-the-art models in short-term load forecasting.
The architecture used in our work is depicted in Figure \ref{fig:tcn}, which is, except for some minor modifications, the network structure detailed in \cite{ConditionalCNN}. In the first layer of the network we process separately the load information and, when available, the exogenous information such as temperature readings. Later the results will be concatenated together and processed by a deep residual network with $L$ layers. Each layer consists of a residual block with 1D dilated causal convolution, a rectified linear unit (ReLU) activation and finally dropout to prevent overfitting. 
The output layer consists of 1x1 convolution which allows the network to output a one-dimensional vector $\vec{y} \in \real^{\inlen}$ having the same dimensionality of the input vector $\vec{x}$.
To approach multi-step forecasting, we adopt a MIMO strategy. 

\subsection{Related Work}
In the short-term load forecasting relevant literature, CNNs have not been studied to a large extent. Indeed, until recently, these models were not considered for any time-series related problem. Still, several works tried to address the topic; in \cite{DeepEnergy} a deep convolutional neural network model named DeepEnergy is presented. The proposed network is inspired by the first architectures used in ImageNet challenge (e.g, \cite{AlexNet}), alternating convolutional and pooling layers, halving the width of the feature map after each step. According to the provided experimental results, DeepEnergy can precisely predict energy load in the next three days outperforming five other machine learning algorithms including LSTM and FNN.
In \cite{MarinoCNN} a CNN is compared to recurrent and feed forward approaches showing promising results on a benchmark dataset.
In \cite{CNN_RNN} a hybrid approach involving both convolutional and recurrent architectures is presented. The authors integrate different input sources and use convolutional layers to extract meaningful features from the historic load while the recurrent network main task is to learn the system's dynamics. The model is evaluated on a large dataset containing hourly loads from a city in North China and is compared with a three-layer feed forward neural network.
A different hybrid approach is presented in \cite{CNN_LSTM}, the authors process the load information in parallel with a CNN and an LSTM. The features generated by the two networks are then used as an input for a final prediction network (fully connected) in charge of forecasting the day-ahead load. 

\section{Performance Assessment}
\label{sec:experiments}
In this section we perform evaluation and assessment of all the presented architectures. The testing is carried out by means of three use cases that are based on two different datasets used as benchmarks. We first introduce the performance metrics that we considered for both network optimization and testing, then describe the datasets that have been used and finally we discuss results.

\subsection{Performance Metrics}
The efficiency of the considered architectures has been measured and quantified using widely adopted error metrics. 
Specifically, we adopted the Root mean squared error (RMSE) and the Mean Absolute Error (MAE):
\begin{equation*}
\text{RMSE} = \sqrt{\frac{1}{N}\sum\limits_{i=0}^{N-1} \frac{1}{\outlen} \sum\limits_{t=0}^{\outlen-1} (\hat{y}_i\t{t} - y_i\t{t})^2}
\label{eq:rmse}
\end{equation*}
\begin{equation*}
\text{MAE} =\frac{1}{N}\sum\limits_{i=0}^{N-1} \frac{1}{\outlen} \sum\limits_{t=0}^{\outlen-1} \mid \hat{y}_i\t{t} - y_i\t{t} \mid
\label{eq:mae2}
\end{equation*}
 where $N$ is the number of input-output pairs provided to the model in the course of testing, $y_i\t{t}$ and $\hat{y}_i\t{t}$ are respectively the real load values and the estimated load values at time $t$ for sample $i$ (i.e., the $i-th$ time window).
 $\langle \cdot \rangle$ is the mean operator, $\Vert \cdot \Vert_2$ is the euclidean L2 norm, while $\Vert \cdot \Vert_1$ is the L1 norm. $\vec{y} \in \real^{\outlen}$ and $\vec{\hat{y}} \in \real^{\outlen}$ are the real load values and the estimated load values for one sample, respectively.
Still, a more intuitive and indicative interpretation of prediction efficiency of the estimators can be expressed by the normalized root mean squared error which, differently from the two above metrics, is independent from the scale of the data:
\begin{equation*}
\text{NRMSE}_{\%} = \frac{\text{RMSE}}{y_{max} - y_{min}} \cdot 100
\label{eq:nrmse}
\end{equation*}
where $y_{max}$ and $y_{min}$ are the maximum and minimum value of training dataset, respectively. 
In order to quantify the proportion of variance in the target that is explained by the forecasting methods we consider also the $\text{R}^2$ index:
\begin{equation*}
\text{R}^2 = \frac{1}{N} \sum\limits_{i=0}^{N-1} (1 - \frac{\text{RSS}}{\text{TSS}}) 
\end{equation*}
\begin{equation*}
\text{RSS} = {\frac{1}{\outlen} \sum\limits_{t}^{\outlen} (\hat{y}_i\t{t} - y_i\t{t})^2} \quad
\text{TSS} = {\frac{1}{\outlen} \sum\limits_{t}^{\outlen} (y_i\t{t} - \bar{y}_i)^2}
\end{equation*}
where $\bar{y}_i = \frac{1}{\outlen} \sum\limits_{t}^{\outlen} y_i\t{t}$

All considered models have been implemented in Keras 2.12 \cite{Keras} with Tensorflow \cite{Tensorflow} as backend. The experiments are executed on a Linux cluster with an Intel(R) Xeon(R) Silver CPU and an Nvidia Titan XP.

\subsection{Use Case I}
The first use case considers the \uci (IHEPC) which contains 2.07M measurements of electric power consumption for a single house located in Sceaux (7km of Paris, France). Measurements are collected every minute between December 2006 and November 2010 (47 months) \cite{UCI}. In this study we focus on prediction of the "Global active power" parameter.
Nearly 1.25\% of measurements are missing, still, all the available ones come with timestamps.
We reconstruct the missing values using the mean power consumption for the corresponding time slot across the different years of measurements. 
In order to have a unified approach we have decided to resample the dataset using a sampling rate of 15 minutes which is a widely adopted standard in modern smart meters technologies. In Table \ref{tab:samples} the sample size are outlined for each dataset.

In this use case we performed the forecasting using only historical load values.  
The right side of Figure \ref{fig:gc_uci} depicts the average weekly electric consumption. 
As expected, it can be observed that the highest consumption is registered in the morning and evening periods of day when the occupancy of resident houses is high. Moreover, the average load profile over a week clearly shows that weekdays are similar while weekends present a different trend of consumption. 

The figure shows that the data are characterized by high variance. 
The prediction task consists in forecasting the electric load for the next day, i.e., 96 timesteps ahead. 


\begin{table}[t]
\centering
    \begin{tabular}{c|c|c}
        \toprule
        Dataset               &\textbf{Train} &  \textbf{Test} \\
        \midrule
        \ucishort             & 103301        & 35040     \\  
        \gcshort              & 44640         & 8928  \\
        \bottomrule
    \end{tabular}
\caption{Sample size of train, validation and test sets for each dataset.}
\label{tab:samples}
\end{table}

In order to assess the performance of the architectures we hold out a portion of the data which denotes our test set and comprises the last year of measurements. The remaining measurements are repeatedly divided in two sets, keeping aside a month of data every five ones. This process allows us to build a training set and a validation set for which different hyper-parameters configurations can be evaluated. Only the best performing configuration is later evaluated on the test set.



\subsection{Use Case II and III}
\begin{figure*}[t]
\begin{multicols}{2}
    \includegraphics[width=\linewidth]{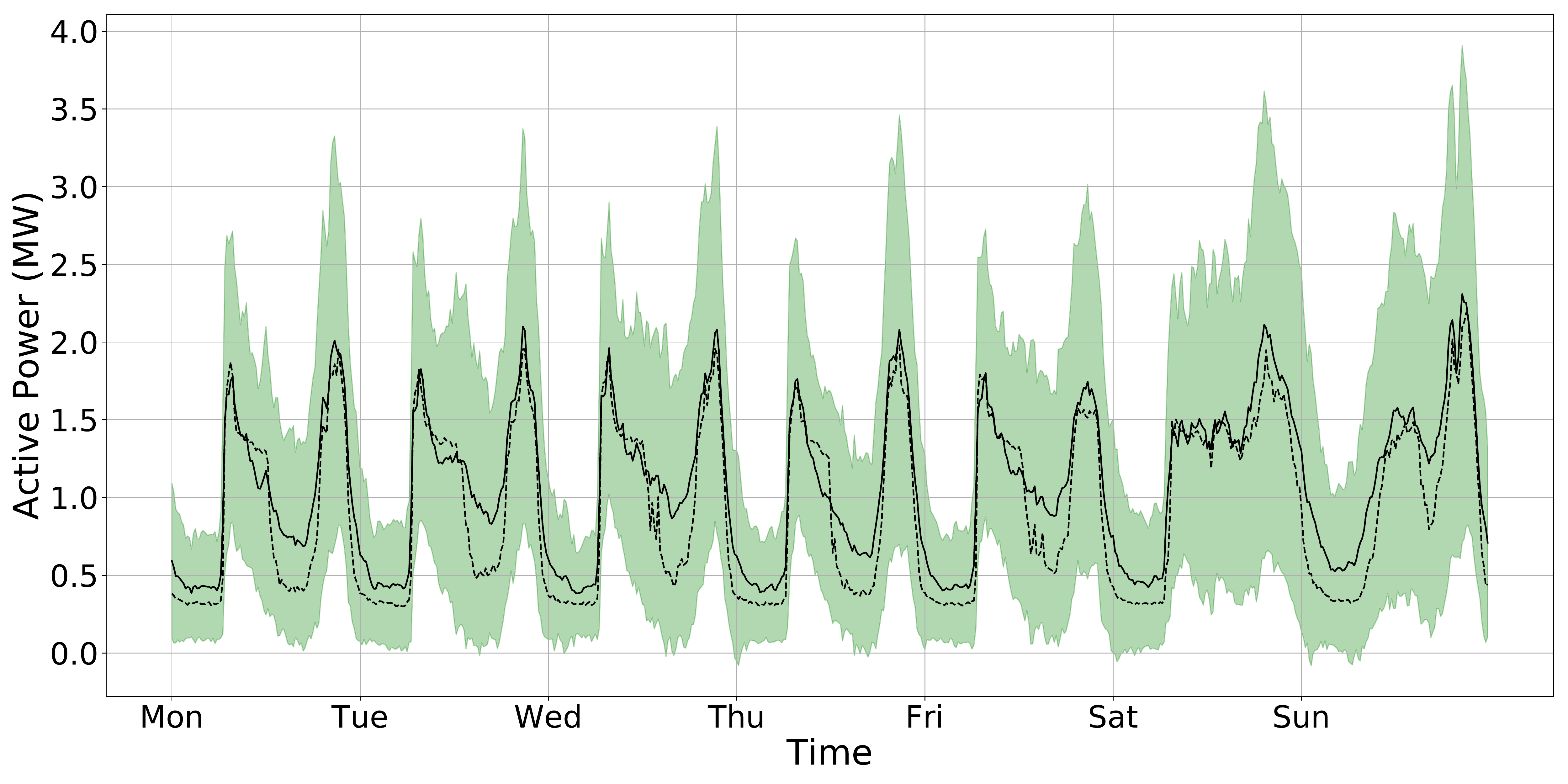}\par
    \includegraphics[width=\linewidth]{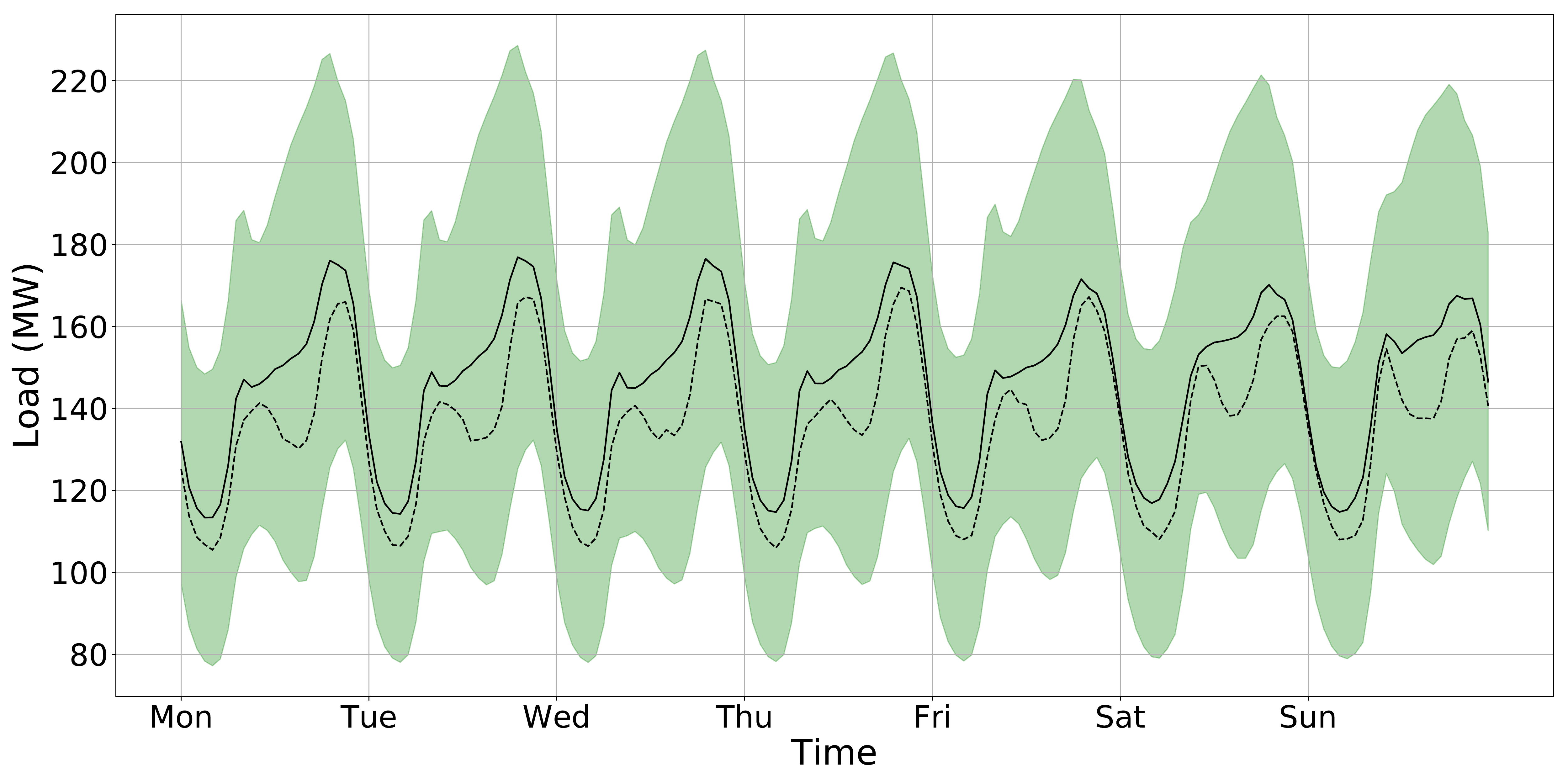}\par 
\end{multicols}
\caption{Weekly statistics for the electric load in the whole \ucishort (Left) and \gcshort datasets (right). The bold line is the mean curve, the dotted line is the median and the green area covers one standard deviation from the mean.}
\label{fig:gc_uci}
\end{figure*}

The other two use cases are based on the \gcshort dataset \cite{GEFCom2014}, which was made available for an online forecasting competition that lasted between August 2015 and December 2015. The dataset contains 60.6k hourly measurements of (aggregated) electric power consumption collected by ISO New England between January 2005 and December 2011. Differently from the \ucishort dataset, temperature values are also available and are used by the different architectures to enhance their prediction performance. In particular the input variables being used for forecasting the subsequent $\outlen$ at timestep $t$ include: several previous load measurements, the temperature measurements for the previous timesteps registered by 25 different stations, hour, day, month and year of the measurements.
We apply standard normalization to load and temperature measurements while for the other variables we simply apply one-hot encoding, i.e., a $K$-dimensional vector in which one of the elements equals 1, and all remaining elements equal 0 \cite{Bishop}.
On the right side of Figure \ref{fig:gc_uci} we observe the average load and the data dispersion on a weekly basis.  Compared to IHEPC, the load profiles look much more regular. This meets intuitive expectations as the load measurements in the first dataset come from a single household, thus the randomness introduced by user behaviour makes more remarkable impact on the results. On the opposite, the load information in \gcshort comes from the aggregation of the data provided by several different smart meters; clustered data exhibits a more stable and regular pattern.
The main task of these use cases, as well the previous one, consists in forecasting the electric load for the next day, i.e., 24 timesteps ahead.
The hyper-parameters optimization and the final score for the models follow the same guidelines provided for IHEPC, the number of points for each subset is described in Table \ref{tab:samples}.



\subsection{Results}
The compared architectures are the ones presented in  previous sections with one exception. We have additionally considered a deeper variant of a feed forward neural network with residual connections which is named DFNN in the remainder of the work. 
In accordance to the findings of \cite{DemystifyingResNet} we have employed a 2-shortcut network, i.e., the input undergoes two affine transformations each followed by a non linearity before being summed to its original values. For regularization purposes we have included Dropout and Batch Normalization \cite{BatchNorm} in each residual block. We have additionally inserted  this model in the results comparison because it represents an evolution of standard feed forward neural networks which is expected to better handle highly complex time-series data.

Table \ref{tab:uci_hyper} summarizes the best configurations found trough grid search for each model and use case.
For both datasets we experimented different input sequences of length $\inlen$. Finally, we used a window size of four days, which represents the best trade-off between performance and memory requirements. The output sequence length $\outlen$ is fixed to one day. 
For each model we identified the optimal number of stacked layers in the network $L$, the number of hidden units per layer $n_H$, the regularization coefficient $\lambda$ (L2 regularization) and the dropout rate $p_d$. 
Moreover, for TCN we additionally tuned the width $k$ of the convolutional kernel and the number of filters applied at each layer $M$ (i.e., the depth of each output volume after the convolution operation). The dilation factor is increased exponentially with the depth of the network, i.e. $d=2^\ell$ with $\ell$ being the $\ell-th$ layer of the network.

\begin{table*}[t]
  \centering
  \def\arraystretch{1.1} 
  \setlength{\tabcolsep}{\hospace} 
  \begin{tabular}{ccccccccccccc}
    \toprule
    \multirow{2}{*}{\textbf{Hyperparameters}} & \multirow{2}{*}{\textbf{Dataset}} & \multirow{2}{*}{\textbf{FNN}} & \multirow{2}{*}{\textbf{DFNN}} & \multirow{2}{*}{\textbf{TCN}} & \multicolumn{2}{c}{\textbf{ERNN}} & \multicolumn{2}{c}{\textbf{LSTM}}           & \multicolumn{2}{c}{\textbf{GRU}} & \multicolumn{2}{c}{\textbf{seq2seq}} \\
    &                                   &                               &                                &                               & \textbf{Rec}    & \textbf{MIMO}   & \textbf{Rec}         & \textbf{MIMO}
    & \textbf{Rec}                      & \textbf{MIMO}                 & \textbf{TF}                    & \textbf{SG}              \\
    \toprule
    \multirow{3}{*}{L}          & \textbf{IHPEC}                  & 3                     & 6      & 8     & 3      & 1      & 2                    & 1                    & 2                    & 1                    & 1                    & 1                    \\
                                & \textbf{GEFCOM}                 & 1                     & 6      & 6     & 4      & 4      & 4                    & 2                    & 4                    & 1                    & 2                    & 1                    \\
                                & \textbf{$\text{GEFCOM}_{exog}$} & 1                     & 6      & 8     & 2      & 1      & 4                    & 1                    & 2                    & 2                    & 2                    & 3                    \\
    \midrule
    \multirow{3}{*}{$n_H$}      & \textbf{IHPEC}                  & $\text{256-128}_{x2}$ & 50     &       & 10     & 30     & 20                   & 20                   & 10                   & 50                   & 30                   & 50                   \\
                                & \textbf{GEFCOM}                 & 60                    & 30     &       & 20     & 50     & 15                   & 20                   & 30                   & 20                   & 10                   & 50                   \\
                                & \textbf{$\text{GEFCOM}_{exog}$} & 60                    & 30     &       & 10     & 30     & 30                   & 50                   & 10                   & 15                   & 20                   & 20-15-10             \\
    \midrule
    \multirow{3}{*}{$\lambda$}  & \textbf{IHPEC}                  & 0.001                 & 0.0005 & 0.005 & 0.001  & 0.001  & 0.001                & 0.001                & 0.001                & 0.0005               & 0.01                 & 0.01                 \\
                                & \textbf{GEFCOM}                 & 0.01                  & 0.0005 & 0.01  & 0.01   & 0.0005 & 0.001                & 0.001                & 0.01                 & 0.0005               & 0.01                 & 0.01                 \\
                                & \textbf{$\text{GEFCOM}_{exog}$} & 0.005                 & 0.0005 & 0.005 & 0.0005 & 0.001  & 0.0005               & 0.0005               & 0.001                & 0.01                 & 0.001                & 0.01                 \\
    \midrule
    \multirow{3}{*}{$p_{drop}$} & \textbf{IHPEC}                  & 0.1                   & 0.1    & 0.1   & 0.0    & 0.0    & 0.0                  & 0.0                  & 0.0                  & 0.0                  & 0.1                  & 0.2                  \\
                                & \textbf{GEFCOM}                 & 0.1                   & 0.1    & 0.1   & 0.1    & 0.0    & 0.1                  & 0.0                  & 0.1                  & 0.0                  & 0.1                  & 0.1                  \\
                                & \textbf{$\text{GEFCOM}_{exog}$} & 0.1                   & 0.1    & 0.1   & 0.0    & 0.0    & 0.0                  & 0.0                  & 0.1                  & 0.0                  & 0.1                  & 0.0                  \\
    \midrule
    \multirow{3}{*}{$k,$ M}     & \textbf{IHPEC}                  &                       &        & 2, 32 &        &        &                      &                      &                      &                      &                      &                      \\
                                & \textbf{GEFCOM}                 &                       &        & 2, 16 &        &        & \multicolumn{1}{l}{} & \multicolumn{1}{l}{} & \multicolumn{1}{l}{} & \multicolumn{1}{l}{} & \multicolumn{1}{l}{} & \multicolumn{1}{l}{} \\
                                & \textbf{$\text{GEFCOM}_{exog}$} &                       &        & 2, 64 &        &        & \multicolumn{1}{l}{} & \multicolumn{1}{l}{} & \multicolumn{1}{l}{} & \multicolumn{1}{l}{} & \multicolumn{1}{l}{} & \multicolumn{1}{l}{} \\
    \bottomrule
  \end{tabular}
  \caption{Best configurations found via Grid Search for the \ucishort dataset, \gcshort dataset and \gcshort with exogenous features.}
  \label{tab:uci_hyper}
\end{table*}

Table \ref{tab:uci_res} summarizes the test scores of the presented architectures obtained for the \ucishort dataset. Certain similarities among networks trained for different uses cases can be spotted out already at this stage. In particular, we observe that all models exploit a small number of neurons. This is not usual in deep learning but - at least for recurrent architectures - is consistent with  \cite{RNN_Bianchi}. With some exceptions, recurrent networks benefit from a less strict regularization; dropout is almost always set to zero and $\lambda$ values are small. 

Among Recurrent Neural Networks we observe that, in general, the MIMO strategy outperforms the recursive one in this multi step prediction task. This is reasonable in such a scenario. Indeed, the recursive strategy, differently from the MIMO one, is highly sensitive to errors accumulation which, in a highly volatile time series as the one addressed here, results in a very inaccurate forecast.
Among the MIMO models we observe that gated networks perform significantly better than simple Elmann network one. This suggests that gated systems are effectively learning to better exploit the temporal dependency in the data.
In general we notice that all the models, except the RNNs trained with recursive strategy, achieve comparable performance and none really stands out. It is interesting to comment that GRU-MIMO and LSTM-MIMO outperform sequence to sequence architectures which are supposed to better model complex temporal dynamics like the one exhibited by the residential load curve. Nevertheless, by observing the performance of recurrent networks trained with the recursive strategy, this  behaviour is less surprising.
In fact, compared with the aggregated load profiles, the load curve belonging to a single smart meter is way more volatile and sensitive to customers behaviour. For this reason, leveraging geographical and socio-economic features that characterize the area where the user lives may allow deep networks to generate better predictions.

\begin{figure*}[t]
\begin{multicols}{2}
    \includegraphics[width=\linewidth]{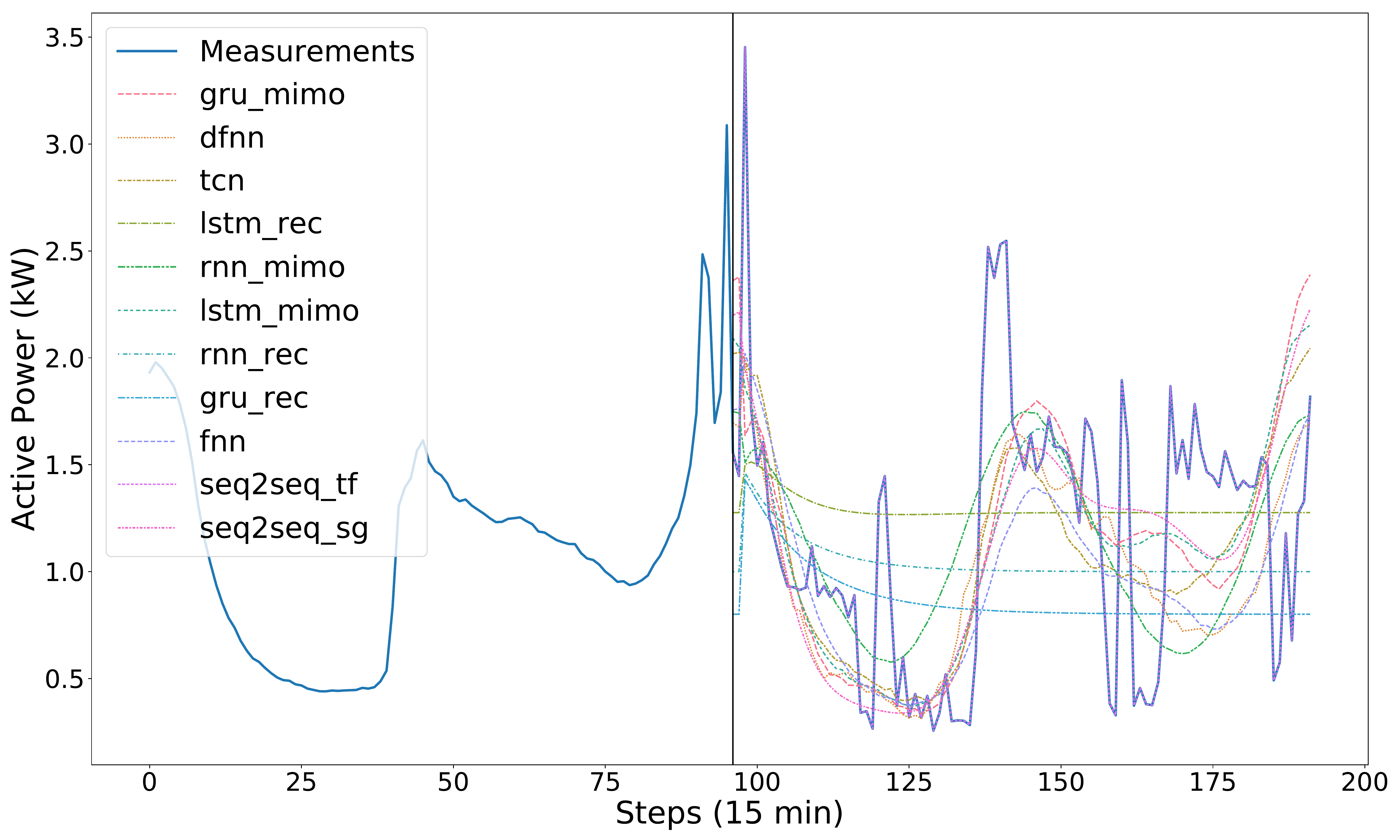}\par
    \includegraphics[width=\linewidth]{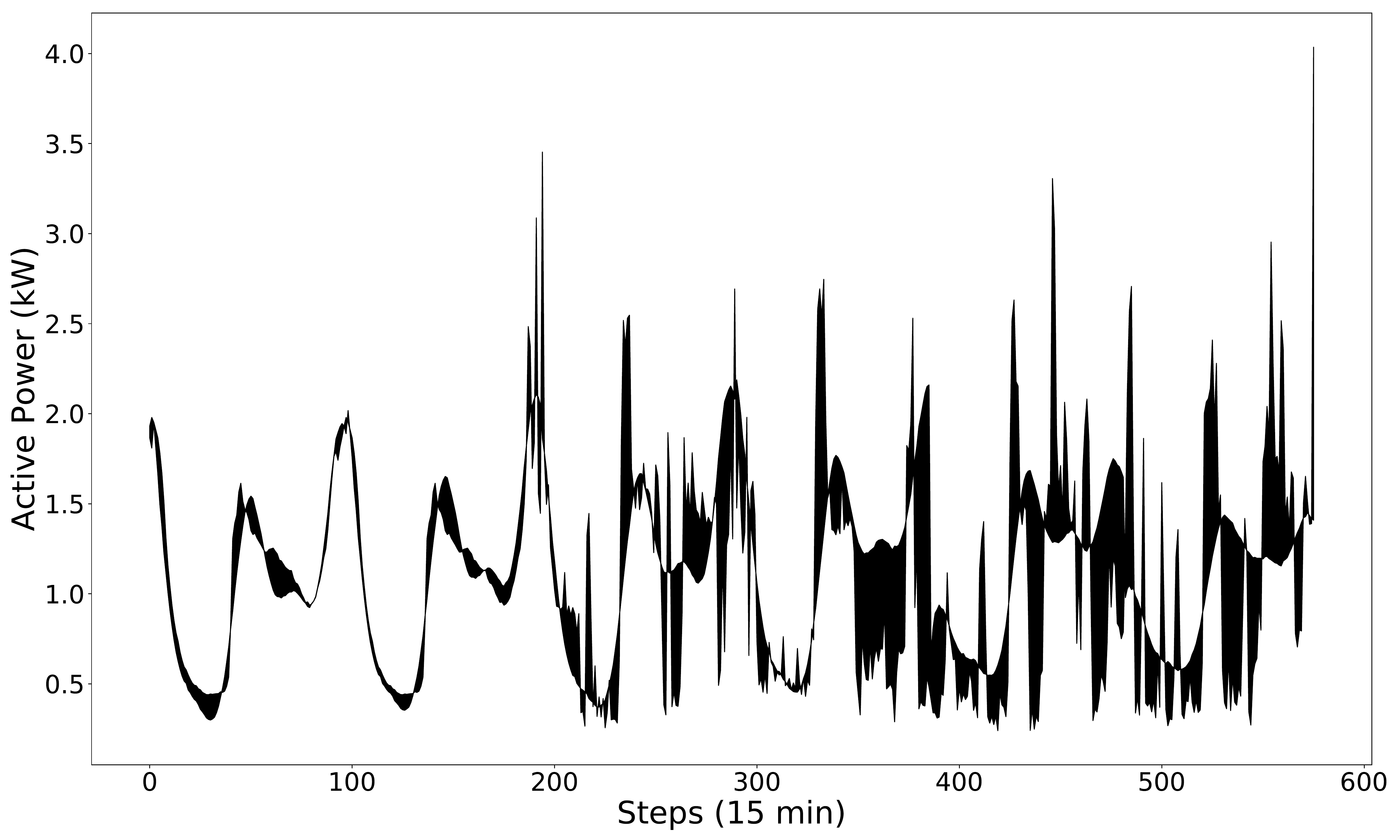}\par
\end{multicols}
\caption{(Right) Predictive performance of all the models on a single day for \ucishort dataset. The left portion of the image shows (part of) the measurements used as input while the right side with multiple lines represents the different predictions. 
(Left) Difference between the best model's predictions (GRU-MIMO) and the actual measurements. The thinner the line the closer the prediction is to the true data.}
\label{fig:uci_pred_plot}
\end{figure*}

For visualization purposes we compare all the models performance on a single day prediction scenario on the left side of Figure \ref{fig:uci_pred_plot}. On the right side of Figure \ref{fig:uci_pred_plot} we quantify the differences between the best predictor (the GRU-MIMO) and the actual measurements; the thinner the line the closer the prediction to the true data.  Furthermore, in this Figure, we concatenate multiple day predictions to have a wider time span and evaluate the model predictive capabilities. We observe that the model is able to generate a prediction that correctly models the general trend of the load curve but fails to predict steep peaks. This might come from the design choice of using MSE as the optimization metric, which could discourage deep models to predict high peaks as large errors are hugely penalized, and therefore, predicting a lower and smoother function results in better performance according to this metric. Alternatively, some of the peaks may simply represent noise due to particular user behaviour and thus unpredictable by definition.

\begin{table}[]
\centering
\def\arraystretch{1.1} 
\setlength{\tabcolsep}{3pt} 
\begin{tabular}{cccccc}
\toprule
                                  &                                 & \textbf{RMSE}        & \textbf{MAE}         & \textbf{NRMSE}    & $\text{R}^2$   \\
\toprule
\multicolumn{2}{c}{\textbf{FNN}}                                & $0.76 \scriptstyle{\pm 0.01}$ & $0.53 \scriptstyle{\pm 0.01}$ & $10.02 \scriptstyle{\pm 0.17}$ & $0.250 \scriptstyle{\pm 0.026}$ \\ \hline
\multicolumn{2}{c}{\textbf{DFNN}}                                & $0.75 \scriptstyle{\pm 0.01}$ & $0.53 \scriptstyle{\pm 0.01}$ & $9.90 \scriptstyle{\pm 0.05}$ & $0.269 \scriptstyle{\pm 0.007}$ \\ \hline
\multicolumn{2}{c}{\textbf{TCN}}    & $0.76 \scriptstyle{\pm 0.008}$ &                    $0.54 \scriptstyle{\pm 0.00}$   & $10.07 \scriptstyle{\pm 0.11}$ & $0.245 \scriptstyle{\pm 0.017}$ \\  \hline
\multirow{2}{*}{\textbf{ERNN}}    & \textbf{MIMO}                     & $0.79 \scriptstyle{\pm 0.00}$                     & $0.56 \scriptstyle{\pm 0.00}$ &                    $10.33 \scriptstyle{\pm 0.08}$   & $0.201 \scriptstyle{\pm 0.012}$ \\ \cline{2-6} 
                                  & \textbf{Rec}                    &$0.88 \scriptstyle{\pm 0.02}$                     & $0.69 \scriptstyle{\pm 0.03}$ &                    $11.61 \scriptstyle{\pm 0.29}$   & $0.001 \scriptstyle{\pm 0.039}$ \\ \hline
\multirow{2}{*}{\textbf{LSTM}}    & \textbf{MIMO}                     & $0.75 \scriptstyle{\pm 0.00}$                     & $0.53 \scriptstyle{\pm 0.00}$ &                    $9.85 \scriptstyle{\pm 0.04}$   & $0.276 \scriptstyle{\pm 0.006}$ \\ \cline{2-6} 
                              & \textbf{Rec}                    & $0.84 \scriptstyle{\pm 0.06}$                     & $0.60 \scriptstyle{\pm 0.07}$ &                    $11.06 \scriptstyle{\pm 0.74}$   & $0.085 \scriptstyle{\pm 0.125}$ \\ \hline
\multirow{2}{*}{\textbf{GRU}}     & \textbf{MIMO}   & $\boldsymbol{0.75 \scriptstyle{\pm 0.00}}$ &                    $\boldsymbol{0.52 \scriptstyle{\pm 0.00}}$   & $\boldsymbol{9.83 \scriptstyle{\pm 0.03}}$ & $\boldsymbol{0.279 \scriptstyle{\pm 0.004}}$\\ \cline{2-6} 
                              & \textbf{Rec}                    & $0.89 \scriptstyle{\pm 0.02}$                     & $0.70 \scriptstyle{\pm 0.02}$ &                    $11.64 \scriptstyle{\pm 0.23}$   & $0.00 \scriptstyle{\pm 0.04}$ \\ \hline
\multirow{2}{*}{\textbf{seq2seq}} & \textbf{TF} & $0.78 \scriptstyle{\pm 0.01}$                    & $0.57 \scriptstyle{\pm 0.02}$                      & $10.22 \scriptstyle{\pm 0.17}$                     & $0.221 \scriptstyle{\pm 0.026}$  \\ \cline{2-6} 
& \textbf{SG} & $0.76 \scriptstyle{\pm 0.01}$                   & $0.53 \scriptstyle{\pm 0.01}$                      & $10.00 \scriptstyle{\pm 0.14}$     &  $0.253 \scriptstyle{\pm 0.03}$               \\ 
\bottomrule
\end{tabular}
\caption{\uci results. Each model's mean score ($\pm$ one standard. deviation) comes from 10 repeated training processes.}
\label{tab:uci_res}
\end{table}

The load curve of the second dataset (\gcshort) results from the aggregation of several different load profiles producing a smoother load curve when compared with the individual load case.
Hyper-parameters optimization and the final score for the models can be found in Table \ref{tab:uci_hyper}.

Table \ref{tab:gefcom_res} and Table \ref{tab:gefcom_res_exog} show the experimental results obtained by the models in two different scenarios. In the former case, only load values were provided to the models while in the latter scenario the input vector has been augmented with the exogenous features described before.
Compared to the previous dataset this time series exhibits a much more regular pattern; as such we expect the prediction task to be easier. Indeed, we can observe a major improvement in terms of performance across all the models.
As already noted in \cite{Residential, STLF_LSTM} the prediction accuracy increases significantly when the forecasting task is carried out on a smooth load curve (resulting from the aggregation of many individual consumers).

We can observe that, in general, all models except plain FNNs benefit from the presence of exogenous variables. 

When exogenous variables are adopted, we notice a major improvement by RNNs trained with the recursive strategy which outperform MIMO ones. This increase in accuracy can be attributed to a better capacity of leveraging the exogenous time series of temperatures to yield a better load forecast. Moreover, RNNs with MIMO strategy gain negligible improvements compared to their performance when no extra-feature is provided. This kind of architectures use a feedforward neural network to map their final hidden state to a sequence of $\outlen$ values, i.e., the estimates. Exogenous variables are elaborated directly by this FNN, which, as observed above, shows to have problems in handling both load data and extra information. Consequently, a better way of injecting exogenous variables in MIMO recurrent network needs to be found in order to provide a boost in prediction performance comparable to the one achieved by employing the recursive strategy. 

For reasons that are similar to those discussed above, sequence to sequence models trained via \textit{teacher forcing} (seq2seq-TF) experienced an improvement when exogenous features are used. Still, seq2seq trained in free-running mode (seq2seq-SG) proves to be a valid alternative to standard seq2seq-TF producing high quality predictions in all use cases. The absence of a discrepancy between training and inference in terms of data generating distribution shows to be an advantage as seq2seq-SG is less sensitive to noise and error propagation.

Finally, we notice that TCNs perform well in all the presented use cases. Considering their lower memory requirements in the training process along with their inherent parallelism this type of networks represents a promising alternative to recurrent neural networks for short-term load forecasting.

\begin{figure*}[t]
\begin{multicols}{2}
    \includegraphics[width=\linewidth]{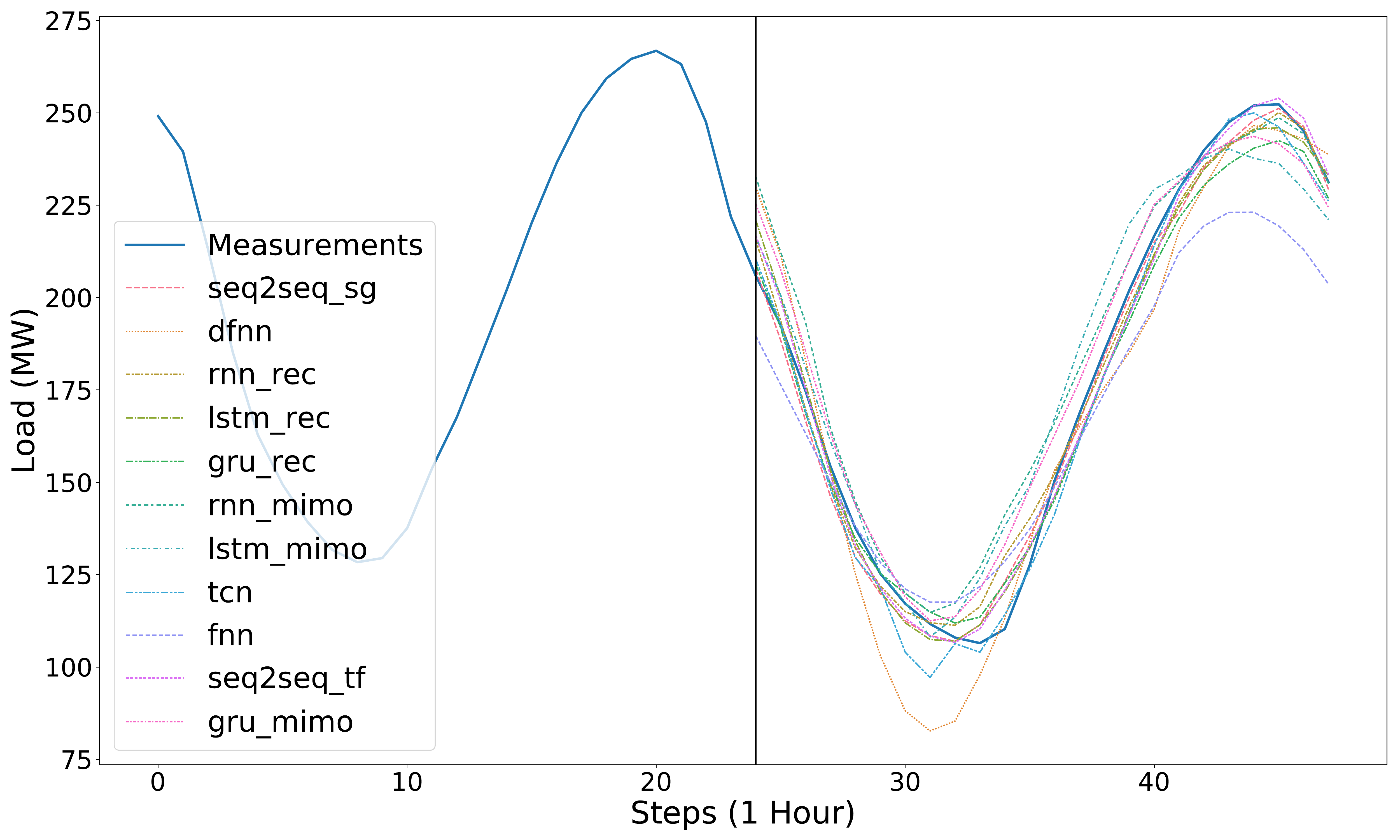}\par
    \includegraphics[width=\linewidth]{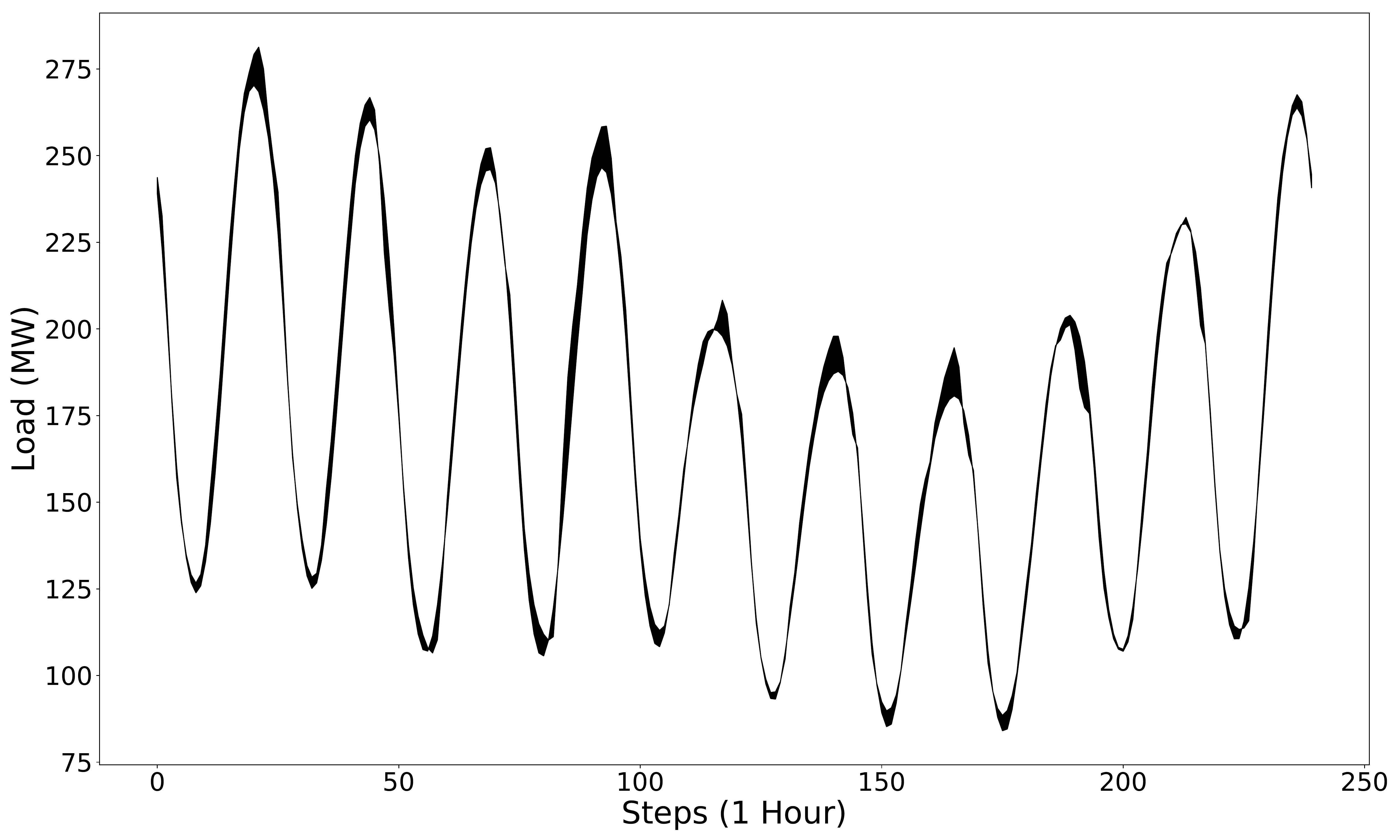}\par
\end{multicols}
\caption{(Right) Predictive performance of all the models on a single day for \gcshort dataset. The left portion of the image shows (part of) the measurements used as input while the right side with multiple lines represents the different predictions. 
(Left) Difference between the best model's predictions (LSTM-Rec) and the actual measurements. The thinner the line the closer the prediction to the true data.}
\label{fig:gefcom_pred_plot}
\end{figure*}

The results of predictions are presented in the same fashion as for the previous use case in Figure \ref{fig:gefcom_pred_plot}. 
Observe that, in general, all the considered models are able to produce reasonable estimates as sudden picks in consumption are smoothed. Therefore, predictors greatly improve their accuracy when predicting day ahead values for the aggregated load curves with respect to individual households scenario.

\begin{table}[t]
\centering
\def\arraystretch{1.1} 
\setlength{\tabcolsep}{3pt} 
\begin{tabular}{cccccc}
\toprule
                                  &                                 & \textbf{RMSE}        & \textbf{MAE}         & \textbf{NRMSE}    & $\text{R}^2$   \\
\toprule
\multicolumn{2}{c}{\textbf{FNN}}                                & $21.1 \scriptstyle{\pm 2.5}$ & $15.5 \scriptstyle{\pm 2.1}$ & $7.01 \scriptstyle{\pm 0.82}$ & $0.833 \scriptstyle{\pm 0.041}$  \\ \hline
\multicolumn{2}{c}{\textbf{DFNN}}                                & $22.4 \scriptstyle{\pm 6.2}$ & $17.1 \scriptstyle{\pm 6.2}$ & $7.44 \scriptstyle{\pm 2.01}$ & $0.801 \scriptstyle{\pm 0.124}$  \\ \hline
\multicolumn{2}{c}{\textbf{TCN}}    & $17.2 \scriptstyle{\pm 0.1}$ &                    $11.5 \scriptstyle{\pm 0.1}$   & $5.71 \scriptstyle{\pm 0.14}$ & $0.891 \scriptstyle{\pm 0.00}$ \\  \hline
\multirow{2}{*}{\textbf{ERNN}}    & \textbf{MIMO}                     & $18.0 \scriptstyle{\pm 0.3}$                     & $11.9 \scriptstyle{\pm 0.4}$ &                    $5.99 \scriptstyle{\pm 0.11}$   & $0.879 \scriptstyle{\pm 0.046}$ \\ \cline{2-6} 
                                  & \textbf{Rec}                    &$27.0 \scriptstyle{\pm 2.3}$                     & $20.7 \scriptstyle{\pm 2.5}$ &                    $8.95 \scriptstyle{\pm 0.78}$   & $0.732 \scriptstyle{\pm 0.046}$ \\ \hline
\multirow{2}{*}{\textbf{LSTM}}    & \textbf{MIMO}                     & $19.5 \scriptstyle{\pm 0.5}$                     & $13.7 \scriptstyle{\pm 0.6}$ &                    $6.47 \scriptstyle{\pm 0.18}$   & $0.861 \scriptstyle{\pm 0.007}$ \\ \cline{2-6} 
                              & \textbf{Rec}                    & $25.6 \scriptstyle{\pm 2.2}$                     & $18.4 \scriptstyle{\pm 1.3}$ &                    $8.52 \scriptstyle{\pm 0.72}$   & $0.757 \scriptstyle{\pm 0.041}$ \\ \hline
\multirow{2}{*}{\textbf{GRU}}     & \textbf{MIMO}                     & $19.0 \scriptstyle{\pm 0.2}$                     & $13.1 \scriptstyle{\pm 0.3}$ &                    $6.29 \scriptstyle{\pm 0.07}$   & $0.868 \scriptstyle{\pm 0.003}$ \\ \cline{2-6} 
                                  & \textbf{Rec}                    & $26.7 \scriptstyle{\pm 3.3}$                     & $19.8 \scriptstyle{\pm 3.1}$ &                    $8.85 \scriptstyle{\pm 1.09}$   & $0.737 \scriptstyle{\pm 0.064}$ \\ \hline
\multirow{2}{*}{\textbf{seq2seq}} & \textbf{TF}  & $21.5 \scriptstyle{\pm 2.1} $                     &    $15.4 \scriptstyle{\pm 1.9} $                  &  $7.13 \scriptstyle{\pm 0.69} $      &  $0.829 \scriptstyle{\pm 0.034} $ \\ \cline{2-6} 
& \textbf{SG} &  $\boldsymbol{17.1 \scriptstyle{\pm 0.2}}$                 & $\boldsymbol{11.3 \scriptstyle{\pm 0.2}}$                     & $\boldsymbol{5.67 \scriptstyle{\pm 0.06}}$     & $\boldsymbol{0.893 \scriptstyle{\pm 0.002}}$                 \\ 
\bottomrule
\end{tabular}
\caption{\gcshort results without any exogenous variable. Each model's mean score ($\pm$ one standard. deviation) comes from 10 repeated training processes.}
\label{tab:gefcom_res}
\end{table}

\begin{table}[t]
\centering
\def\arraystretch{1.1} 
\setlength{\tabcolsep}{3pt} 
\begin{tabular}{cccccc}
\toprule
                                  &                                 & \textbf{RMSE}        & \textbf{MAE}         & \textbf{NRMSE}    & $\text{R}^2$   \\
\toprule
\multicolumn{2}{c}{\textbf{FNN}}                                & $27.9 \scriptstyle{\pm 2.8}$ & $20.8 \scriptstyle{\pm 2.4}$ & $9.28 \scriptstyle{\pm 0.93}$ & $0.709 \scriptstyle{\pm 0.062}$  \\ \hline
\multicolumn{2}{c}{\textbf{DFNN}}                                & $23.0 \scriptstyle{\pm 1.2}$ & $15.6 \scriptstyle{\pm 0.7}$ & $7.62 \scriptstyle{\pm 0.41}$ & $0.805 \scriptstyle{\pm 0.021}$  \\ \hline
\multicolumn{2}{c}{\textbf{TCN}}    & $15.4 \scriptstyle{\pm 1.5}$ &                    $8.6 \scriptstyle{\pm 1.7}$   & $5.00 \scriptstyle{\pm 0.22}$ & $0.917 \scriptstyle{\pm 0.007}$ \\  \hline
\multirow{2}{*}{\textbf{ERNN}}    & \textbf{MIMO}                     & $17.9 \scriptstyle{\pm 0.3}$                     & $11.7 \scriptstyle{\pm 0.3}$ &                    $5.94 \scriptstyle{\pm 0.01}$   & $0.883 \scriptstyle{\pm 0.004}$ \\ \cline{2-6} 
                                  & \textbf{Rec}                    &$14.7 \scriptstyle{\pm 1.0}$                     & $8.6 \scriptstyle{\pm 1.0}$ &                    $4.88 \scriptstyle{\pm 0.19}$   & $0.925 \scriptstyle{\pm 0.005}$ \\ \hline
\multirow{2}{*}{\textbf{LSTM}}    & \textbf{MIMO}                     & $18.1 \scriptstyle{\pm 1.3}$                     & $12.1 \scriptstyle{\pm 1.3}$ &                    $6.01 \scriptstyle{\pm 0.42}$   & $0.877 \scriptstyle{\pm 0.018}$ \\ \cline{2-6} 
                              & \textbf{Rec}                    & $\boldsymbol{13.8 \scriptstyle{\pm 0.6}}$                     & $\boldsymbol{7.5 \scriptstyle{\pm 0.3}}$ &                    $\boldsymbol{4.59 \scriptstyle{\pm 0.18}}$   & $\boldsymbol{0.930 \scriptstyle{\pm 0.006}}$ \\ \hline
\multirow{2}{*}{\textbf{GRU}}     & \textbf{MIMO}                     & $17.8 \scriptstyle{\pm 0.2}$                     & $11.7 \scriptstyle{\pm 0.2}$ &                    $5.93 \scriptstyle{\pm 0.07}$   & $0.882 \scriptstyle{\pm 0.002}$ \\ \cline{2-6} 
                                  & \textbf{Rec}                    & $16.7 \scriptstyle{\pm 0.5}$                     & $10.0 \scriptstyle{\pm 0.6}$ &                    $5.54 \scriptstyle{\pm 0.15}$   & $0.898 \scriptstyle{\pm 0.006}$ \\ \hline
\multirow{2}{*}{\textbf{seq2seq}} & \textbf{TF} & $14.3 \scriptstyle{\pm 1.0}$                    & $8.5 \scriptstyle{\pm 0.9}$                     & $4.74 \scriptstyle{\pm 0.32}$                      & $0.924 \scriptstyle{\pm 0.014}$        \\ \cline{2-6} 
& \textbf{SG} &  $15.9 \scriptstyle{\pm 1.8}$                 & $9.8 \scriptstyle{\pm 1.8}$                     & $5.28 \scriptstyle{\pm 0.60}$     & $0.907 \scriptstyle{\pm 0.021}$                 \\ 
\bottomrule
\end{tabular}
\caption{\gcshort results with exogenous variables. Each model's mean score ($\pm$ one standard. deviation) comes from 10 repeated training processes.}
\label{tab:gefcom_res_exog}
\end{table}

\section{Conclusions}
\label{sec:conclusions}
In this work we have surveyed and experimentally evaluated the most relevant deep learning models applied to the short-term load forecasting problem, paving the way for standardized assessment and identification of the most optimal solutions in this field. The focus has been given to the three main families of models, namely, Recurrent Neural Networks, Sequence to Sequence Architectures and recently developed Temporal Convolutional Neural Networks. An architectural description along with a technical discussion on how multi-step ahead forecasting is achieved, has been provided for each considered model. Moreover, different forecasting strategies are discussed and evaluated, identifying advantages and drawbacks for each of them. The evaluation has been carried out on the three real-world use cases that refer to two distinct scenarios for load forecasting. Indeed, one use case deals with dataset coming from a single household while the other two tackle the prediction of a load curve that represents several aggregated meters, dispersed over the wide area.
Our findings concerning application of recurrent neural networks to short-term load forecasting, show that the simple ERNN performs comparably to gated networks such as GRU and LSTM when adopted in aggregated load forecasting. Thus, the less costly alternative provided by ERNN may represent the most effective solution in this scenario as it allows to reduce the training time without remarkable impact on prediction accuracy.
On the contrary, a significant difference exists for single house electric load forecasting where the gated networks shows to be superior to Elmann ones suggesting that the gated mechanism allows to better handle irregular time series.
Sequence to Sequence models have demonstrated to be quite efficient in load forecasting tasks even though they seem to fail in outperforming RNNs. In general we can claim that seq2seq architectures do not represent a golden standard in load forecasting as they are in other domains like natural language processing. In addition to that, regarding this family of architectures, we have observed that teacher forcing may not represent the best solution for training seq2seq models on short-term load forecasting tasks. Despite being harder in terms of convergence, free-running models learn to handle their own errors, avoiding the discrepancy between training and testing that is a well known issue for teacher forcing. It turns out to be worth efforts to further investigate capabilities of seq2seq models trained with intermediate solutions such as \textit{professor forcing}.
Finally, we evaluated the recently developed Temporal Convolutional Neural Networks which demonstrated convincing performance when applied to load forecasting tasks. Therefore, we strongly believe that the adoption of these networks for sequence modelling in the considered field is very promising and might even introduce a significant advance in this area that is emerging as a key importance for future Smart Grid developments. 

\section*{Acknowledgment}
This project is carried out within the frame of the Swiss Centre for Competence in Energy Research on the Future Swiss Electrical Infrastructure (SCCER-FURIES) with the financial support of the Swiss Innovation Agency (Innosuisse - SCCER program).

\bibliographystyle{unsrt}  
\bibliography{bib/bibliography}

\end{document}